\documentclass[letterpaper]{article} 
\usepackage{aaai24}  
\usepackage{times}  
\usepackage{helvet}  
\usepackage{courier}  
\usepackage[hyphens]{url}  
\usepackage{graphicx} 
\usepackage{natbib}  
\usepackage{caption} 
\usepackage{algorithm}
\usepackage{algorithmic}

\usepackage{newfloat}
\usepackage{listings}
\usepackage{amssymb}
\usepackage{booktabs}
\usepackage{amsmath}
\usepackage{multirow}
\usepackage{bbding}
\usepackage{array}
\DeclareCaptionStyle{ruled}{labelfont=normalfont,labelsep=colon,strut=off} 
\floatstyle{ruled}
\newfloat{listing}{tb}{lst}{}
\floatname{listing}{Listing}
\title{HyperEditor: Achieving Both Authenticity and Cross-Domain Capability\\ in Image Editing via Hypernetworks}
\author{
    Hai Zhang\textsuperscript{\rm 1,\rm 2}, Chunwei Wu\textsuperscript{\rm 1,\rm 2}, Guitao Cao\textsuperscript{\rm 1,\rm 2}\thanks{Corresponding author.}, Hailing Wang\textsuperscript{\rm 1,\rm 2}, Wenming Cao\textsuperscript{\rm 3}\\
}
\affiliations{
    \textsuperscript{\rm 1}Shanghai Key Laboratory of Trustworthy Computing, East China Normal University, China\\

    \textsuperscript{\rm 2}MoE Engineering Research Center of SW/HW Co-design Technology and Application, East China Normal University, China\\
    \textsuperscript{\rm 3}College of Information Engineering, Shenzhen University, China\\
    
    \{hzhang22, 52215902005\}@stu.ecnu.edu.cn, gtcao@sei.ecnu.edu.cn, 52215902004@stu.ecnu.edu.cn, wmcao@szu.edu.cn
}

\usepackage{bibentry}

\begin{document}

\maketitle

\begin{abstract}
Editing real images authentically while also achieving cross-domain editing remains a challenge. Recent studies have focused on converting real images into latent codes and accomplishing image editing by manipulating these codes. However, merely manipulating the latent codes would constrain the edited images to the generator's image domain, hindering the attainment of diverse editing goals. In response, we propose an innovative image editing method called HyperEditor, which utilizes weight factors generated by hypernetworks to reassign the weights of the pre-trained StyleGAN2's generator. Guided by CLIP's cross-modal image-text semantic alignment, this innovative approach enables us to simultaneously accomplish authentic attribute editing and cross-domain style transfer, a capability not realized in previous methods. Additionally, we ascertain that modifying only the weights of specific layers in the generator can yield an equivalent editing result. Therefore, we introduce an adaptive layer selector, enabling our hypernetworks to autonomously identify the layers requiring output weight factors, which can further improve our hypernetworks' efficiency. Extensive experiments on abundant challenging datasets demonstrate the effectiveness of our method.
\end{abstract}

\section{Introduction}

The primary objective of image editing is to modify specific properties of real images by leveraging conditions. In recent years, image editing has emerged as one of the most dynamic and vibrant research areas in academia and industry. Several studies \cite{Shen2019InterpretingTL, harkonen2020ganspace, liu2023fine, revanur2023coralstyleclip, liu2023delving, patashnik2021styleclip} have investigated the extensive latent semantic representations in StyleGAN2 \cite{karras2020analyzing} and have successfully achieved diverse and authentic image editing through the manipulation of latent codes. These methods share a common characteristic: finding the optimal target latent codes for substituting the initial latent codes, thereby advancing the source image to the target image. However, latent codes with high reconstructability often exhibit weak editability \cite{pehlivan2023styleres}. Additionally, if the target image falls outside the image domain of the generator, it becomes difficult to achieve cross-domain image editing purely through the manipulation of the latent codes. Thus, we pondered the question:  \textit{“Is it possible to achieve both authentic image attribute editing and cross-domain image style transfer simultaneously?”}

Recently, \cite{alaluf2022hyperstyle,dinh2022hyperinverter} utilize hypernetworks to reassign the generator's weights, achieving a more precise image reconstruction by gradually compensating for the missing details of the source image in the reconstructed image. Inspired by this, we find that modulating the generator's weights can achieve detailed attribute changes in images. Therefore, we adopt the concept of weight reassignment to the image editing task. In our work, we propose a novel image editing method called HyperEditor, which directly conducts image editing by utilizing weight factors generated by hypernetworks to reassign the weights of StyleGAN2's generator. Unlike traditional methods of model weight fine-tuning \cite{gal2022stylegan}, which often involve retraining the pre-trained model, our approach involves scaling and reassigning the weights of the generator using weight factors. As a result, our method offers better controllability, allowing for authentic attribute editing (e.g., facial features, hair, etc.). Moreover, due to our method's capability to modify the generator's weights, it can effectively perform cross-domain editing operations, which might be challenging to achieve solely by manipulating the latent codes. To the best of our knowledge, our method is the first to achieve authentic attribute editing and cross-domain style editing simultaneously.

We pondered a question: must we reassign all layers in StyleGAN2's generator when editing a single attribute? Based on experimental findings, we observed that only a few layers' weight factors undergo significant changes before and after editing. Thus, we propose the adaptive layer selector, enabling hypernetworks to choose the layers that require outputting weight factors autonomously. Consequently, we can maximize the effect of weight factors while achieving comparable results.

\begin{figure*}
    \centering
    \includegraphics[width=0.85\textwidth]{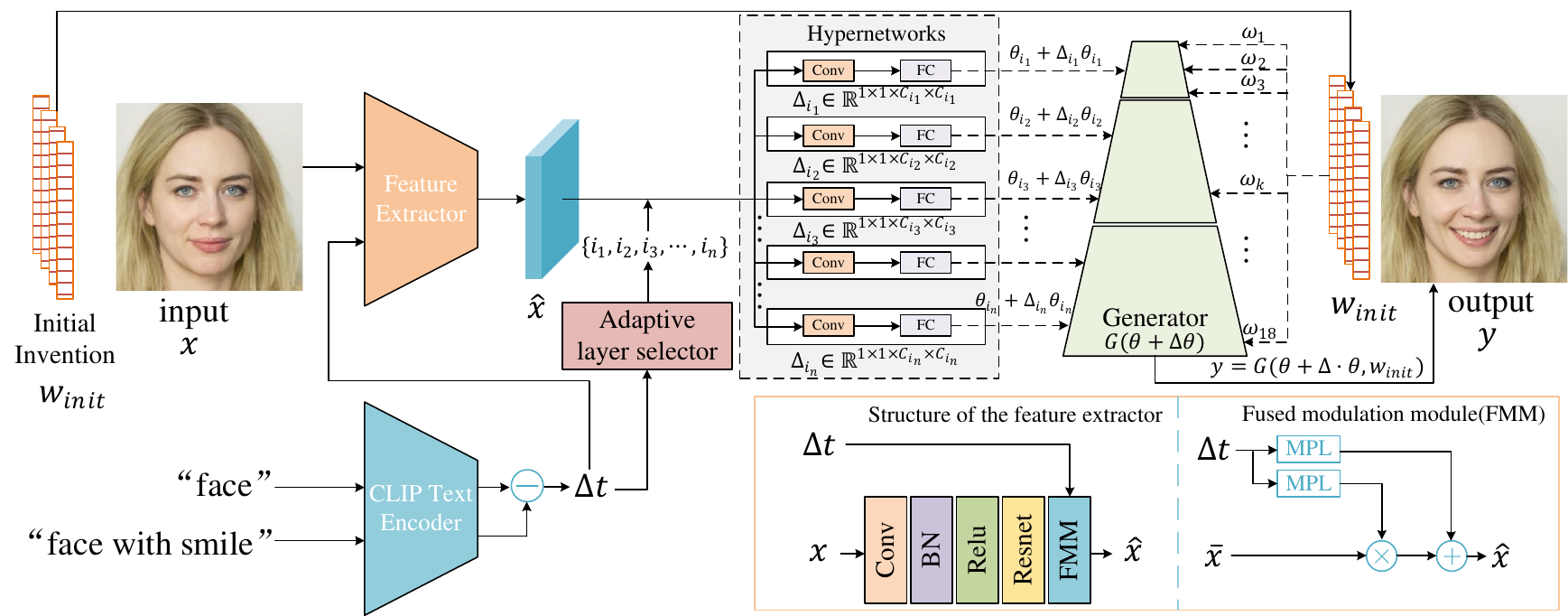}
    \caption{The overall structure of our HyperEditor. Given a text pair $(t_0,t_1)$ and an initial image $x$, we utilize the CLIP text encoder to extract features for $(t_0,t_1)$ and compute their difference, resulting in $\Delta t$. The image feature extractor processes $x$ to obtain its feature representation, which is then refined with the conditional information $\Delta t$ through the fusion modulation module(FMM), yielding an intermediate feature map, $\hat{x}$. Using the adaptive layer selector, we identify a sequence of layers, $L$, that require outputting weight factors. Subsequently, our hypernetworks generate weight factors, $\Delta$, based on $\hat{x}$ and $L$, which are used to reassign the generator's weights. Moreover, we input the latent codes $w_{init}$ of the initial image. Finally, the generated image after editing, $y=(\theta+\Delta \cdot \theta, w_{init})$, is obtained.}
    \label{fig:HyperEditor}
    
\end{figure*}

During model training, aligning the generated images with the target conditions poses a challenge due to the lack of paired datasets before and after editing in the real-world. Recently, some methods \cite{wei2022hairclip, kocasari2022stylemc} use CLIP \cite{radford2021learning} to convert image features into pseudo-text features and align them with genuine-text features. In our work, we leverage the self-supervised learning capacity of CLIP and introduce the directional CLIP loss to supervise model training. This process aligns the difference set of pseudo-text features before and after editing with the difference set of authentic-text pair features, all in a direction-based manner. It makes our approach more focused on cross-modal representations of local semantics, enhances the convergence capability of the model, and prevents the generation of adversarial effects. Additionally, we introduce a fusion modulation module that refines intermediate feature maps using text prompts as scaling and shifting factors. This enables different text prompts to manipulate the hypernetworks and generate various weight factors. Consequently, our approach enables a single model to accomplish diverse image editing tasks. Overall, our contributions can be summarized as follows:

\begin{itemize}
    \item We surpass the constraints of prior image editing techniques and introduce a novel image editing framework called HyperEditor. This framework utilizes hypernetworks to reassign the weights of StyleGAN2's generator and leverages CLIP's cross-modal semantic alignment ability to supervise our model's training. Consequently, HyperEditor can not only authentically modify attributes in images but also achieve cross-domain style editing.
    \item We propose an adaptive layer selector that allows hypernetworks to autonomously determine the layers that require outputting weight factors when editing a single attribute, maximizing the efficiency of the hypernetworks.
\end{itemize}

\section{Related work}
\subsection{Image editing}
Many studies \cite{saha2021loho, zhu2022one, roich2022pivotal} have investigated how to leverage the latent space of pre-trained generators for image editing. Particularly, with the advent of StyleGAN2, image editing through manipulation of the latent codes has become a prevalent research topic. In \cite{Shen2019InterpretingTL}, the authors utilized linear variation to achieve a disentangled representation of the latent codes in StyleGAN. They completed accurate image editing by decoupling entangled semantics and subspace projection. In \cite{harkonen2020ganspace}, the authors proposed using principal component analysis to identify and modify the latent directions in latent codes, thereby enabling image editing. TediGAN \cite{xia2021tedigan} introduced a visual-language similarity module that maps linguistic representations into a latent space that aligns with visual representations, allowing text-guided image editing. StyleCLIP \cite{patashnik2021styleclip} combines the robust cross-modal semantic alignment capability of CLIP with the generative power of StyleGAN2. It presents three text-driven image editing methods, namely latent optimization, latent mapping, and global directions, to achieve image editing in an unsupervised or self-supervised manner. Subsequently, a series of CLIP+StyleGAN2 methods \cite{wei2022hairclip, lyu2023deltaedit} have been introduced. These methods utilize mappers to learn style residuals and transform the original image's latent codes toward the target latent codes. Furthermore, several studies \cite{zeng2022sketchedit, revanur2023coralstyleclip, hou2022feat} have incorporated mask graphs into the network to provide more accurate supervision for specific attribute changes. We can observe that all these methods involve the manipulation of latent codes. However, if the target image exceeds the image domain of the generator, it becomes challenging to rely solely on modifying latent codes to achieve cross-domain editing operations. Thus, we take an approach, focusing on the generator's weights and performing image editing by reassigning these weights. Our method is entirely distinct from theirs, as it does not involve manipulating latent codes during image editing. Our approach sets the groundwork for developing novel image editing techniques.

\begin{figure}
    \centering
    \includegraphics[width=\linewidth]{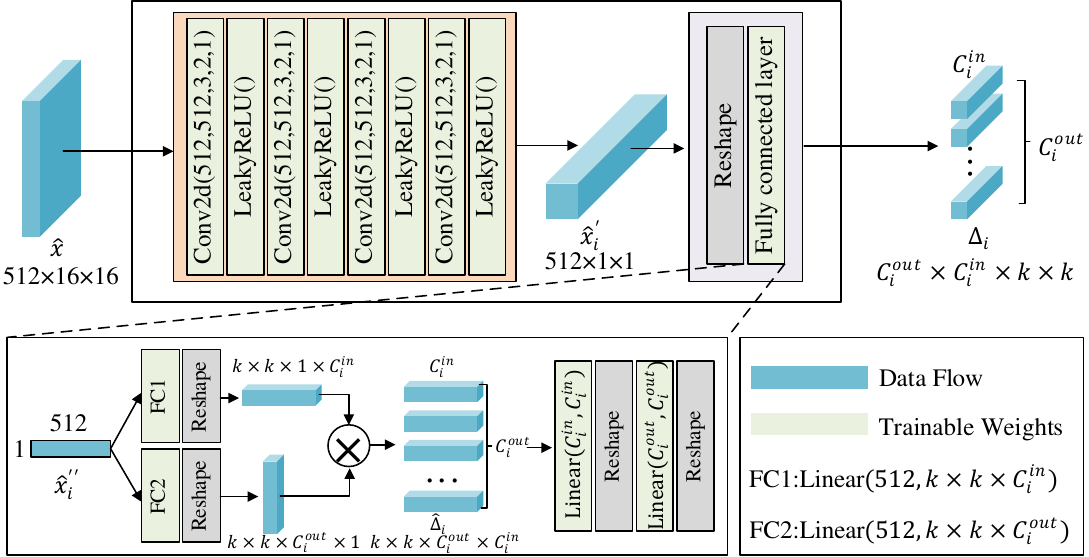}
    \caption{The structure of a hypernetwork. It consists of downsampled convolutions and fully connected layers. The hypernetwork takes the intermediate feature map $\hat{x}$ as input, which predicts and outputs the weight factor $\Delta_i$ for the convolutional layer $i$ in StyleGAN2’s generator.}
    \label{fig:hypernetworks}
    
\end{figure}

\subsection{Hypernetworks}
A hypernetwork is an auxiliary neural network responsible for generating weights for another network, often called the primary network. It was initially introduced by \cite{ha2017hypernetworks}. Training the hypernetworks on extensive datasets can adjust the weights of the main network through appropriate weights shifts, resulting in a more expressive model. Since its proposal, hypernetworks have found applications in various domains, including semantic segmentation \cite{nirkin2021hyperseg}, neural architecture search \cite{zhang2018graph}, 3D modeling \cite{littwin2019deep, sitzmann2020implicit}, continuous learning \cite{Oswald2020Continual}, and more. Recently, hypernetworks have also been applied to the StyleGAN inversion task. Both \cite{alaluf2022hyperstyle} and \cite{dinh2022hyperinverter} have developed hypernetworks structures to enhance the quality of image reconstruction. However, they employ additional methods (such as StyleCLIP, etc.) to complete image editing. In contrast, we utilize hypernetworks directly to accomplish image editing tasks. Moreover, our method can achieve both authentic attribute editing and cross-domain style editing simultaneously. To the best of our knowledge, this is the first occurrence of such an approach in the field of image editing.

\begin{figure}
    \centering
    \includegraphics[width=0.85\linewidth]{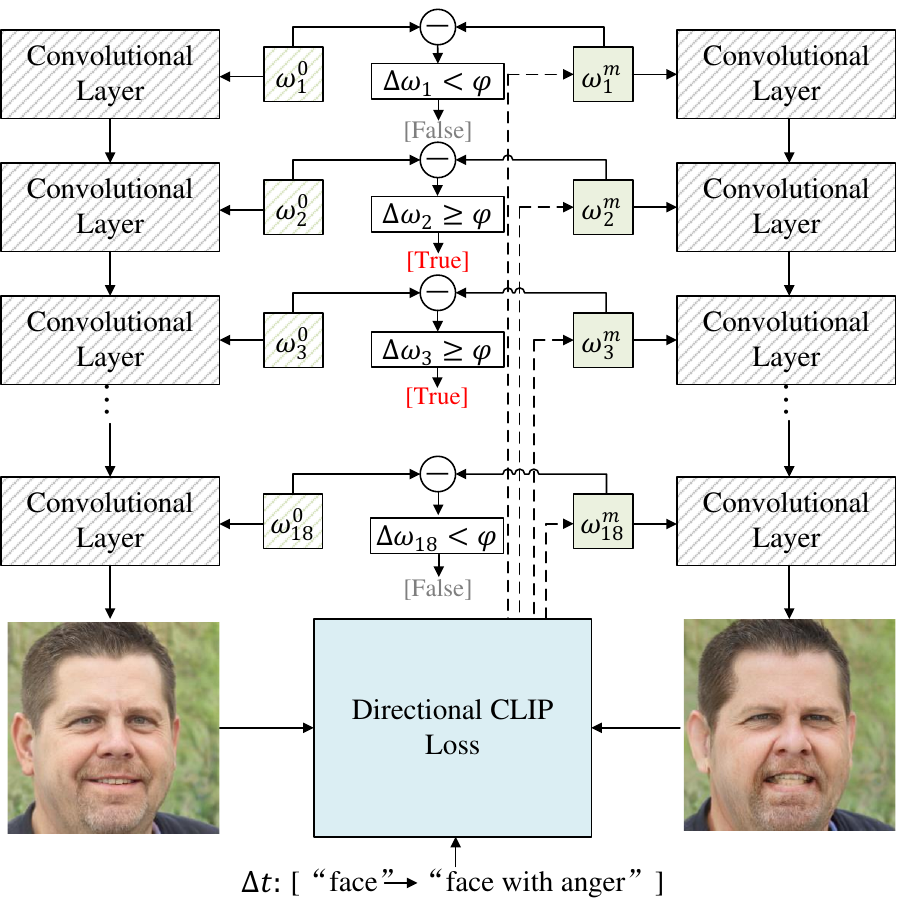}
    \caption{Overview of the adaptive layer selector. We utilize random latent codes as optimization objects and employ directional CLIP loss for a few iterations to dynamically select layers with significant differences between target and source codes. The grid part indicates parameter freezing, while the solid-colored part indicates optimization training.}
    \label{fig:adapt_layer_choose}
    
\end{figure}

\section{Method}
The overall structure of our model is illustrated in Figure \ref{fig:HyperEditor}, providing an overview of the image editing process. The following sections will comprehensively analyze each component in our approach.

\begin{figure*}
    \centering
    \includegraphics[width=0.8\textwidth]{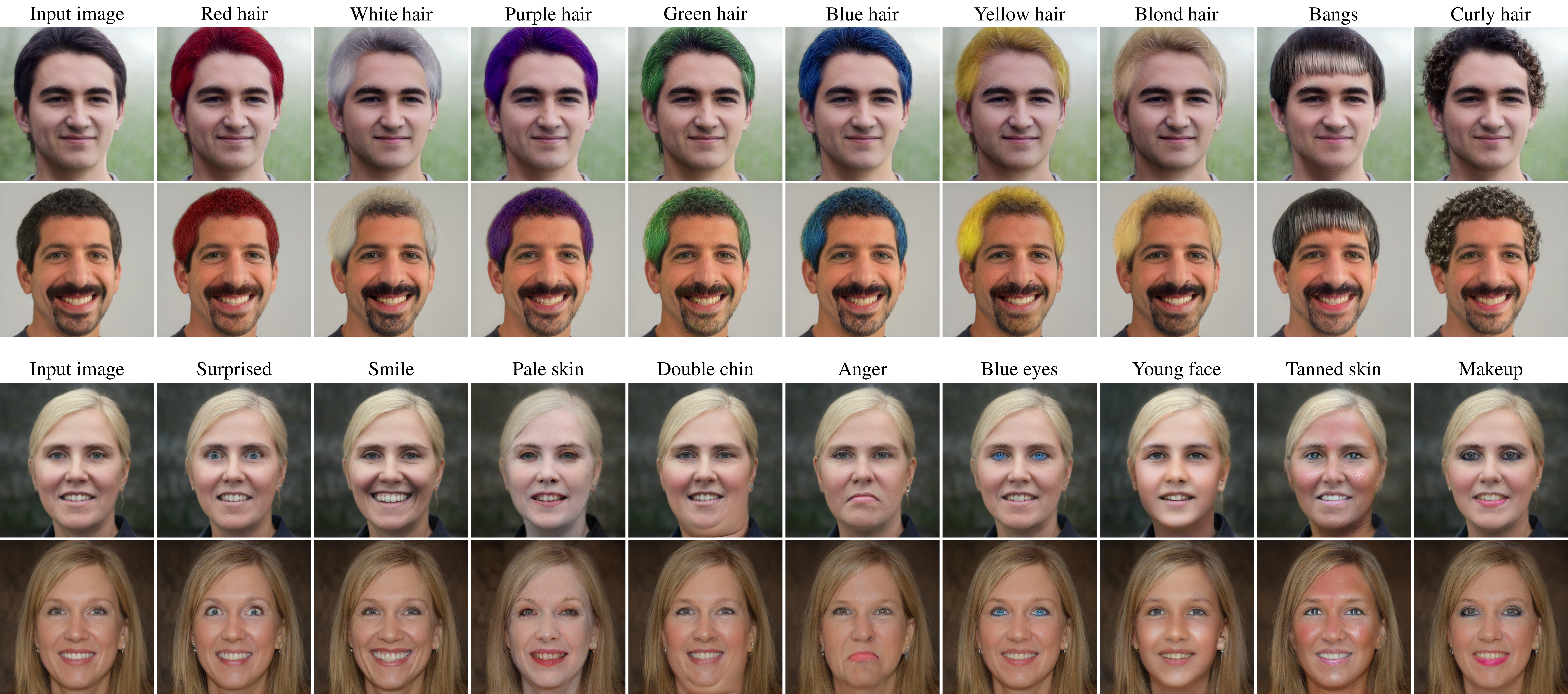}
    \caption{Results of image editing on FFHQ dataset by using our method. The target attributes are located above the images.}
    \label{fig:ffhq_edit}
    
\end{figure*}

\subsection{The design of expressive hypernetworks}
Our approach is to reassign the generator's weights for image editing, which requires our hypernetworks to be expressive, allowing us to control the generator effectively. This control empowers us to edit images both authentically and cross-domain. The details of our hypernetworks are depicted in Figure \ref{fig:hypernetworks}. It takes the intermediate feature map $\hat{x} \in \mathbb{R}^{512\times 16\times 16}$ as input. Firstly, we extract the features of $\hat{x}$ and obtain $\hat{x}_i^{\prime} \in \mathbb{R}^{512 \times 1 \times 1}$ through four down-sampling convolution operations. We then reshape $\hat{x}_i^{\prime}$ as $\hat{x}_i^{\prime \prime} \in \mathbb{R}^{512}$, and it undergoes a series of fully connected operations. In the fully connected operations, we first employ two distinct fully connected layers to expand $\hat{x}_i^{\prime \prime}$ dimensions, yielding two different vectors: $\sigma_1(\hat{x}_i^{\prime \prime}) \in \mathbb{R}^{k \times k \times c_i^{in}}$ and $\sigma_2(\hat{x}_i^{\prime \prime}) \in \mathbb{R}^{k \times k \times c_i^{out}}$. Here, $\sigma_1$ and $\sigma_2$ represent the FC1 and FC2, respectively. Where $C_i^{in}$ and $C_i^{out}$ represent the number of channels per convolution kernel and the total number of convolution kernels in the $i^{th}$ layer of StyleGAN2's generator, respectively. Then, we reshape them and calculate the inner product to obtain the vector $\widehat{\Delta}_i \in \mathbb{R}^{k \times k \times C_i^{out} \times C_i^{in}}$, which is as follows:

\begin{equation}
\widehat{\Delta}_i=R\left(\sigma_1\left(\hat{x}_i^{\prime \prime}\right)\right) \otimes R\left(\sigma_2\left(\hat{x}_i^{\prime \prime}\right)\right)
\end{equation}

Where $R$ denotes the reshape operation, and $\otimes$ represents multidimensional matrix multiplication. To expand the representation space of $\hat{\Delta}_i$, we conduct two consecutive fully connected (FC) operations on $\hat{\Delta}_i$, followed by reshaping, resulting in the weight factors $\Delta_i \in \mathbb{R}^{C_i^{out} \times C_i^{in} \times k \times k}$. We denote the weights of the $i^{th}$ layer in StyleGAN2 as $\theta_i=\{\theta_i^{j, k} \mid 0 \leq j \leq C_i^{out}, 0 \leq k \leq C_i^{in}\}$, where $j$ represents the $j^{th}$ convolutional kernel, and $k$ denotes the $k^{th}$ channel of the convolutional kernel. To achieve image editing, we reassign the weights of the generator based on equation \ref{equ:5}. At this point, we obtain the final generated image after editing $y = G(\hat{\theta}, w_{init})$, where $w_{init}$ is the latent vector of the original image $x$.

\begin{equation}
\hat{\theta}_i^{j, k}=\theta_i^{j, k}+\Delta_i^{j, k} \cdot \theta_i^{j, k}
\label{equ:5}
\end{equation}

\subsection{Directional CLIP guidance}
In the real world, pairwise image datasets where specific attributes change are scarce. It poses a challenge to supervise the alignment of generated images with target images efficiently. To tackle this issue, we leverage the CLIP model's potent cross-modal semantic alignment capability to facilitate the transformation of original images into target images in a self-supervised learning manner. In other words, the training of the entire model can be accomplished using only the original image and the target text prompt. In the existing work, \cite{wei2022hairclip} utilizes CLIP to directly align the generated image with the text conditions, as depicted in equation \ref{formula:1}, where $E_i$ represents the image encoder of CLIP, $E_t$ represents the text encoder of CLIP, $cos(\cdot, \cdot)$ represents the cosine similarity, and $y$ is the generated image.

\begin{equation}
    \mathcal{L}_{{CLIP }}^{{globe }}=1-\cos \left(E_i(y), E_t({ Text })\right)
    \label{formula:1}
\end{equation}

Since specific attribute changes typically involve localized modifications, direct semantic alignment may lead to global feature alterations, making it challenging for the network to converge quickly. To address this issue, we draw inspiration from the approaches in \cite{kwon2022clipstyler, lyu2023deltaedit} and introduce the directional CLIP loss to align the text condition with the image. The process of semantic alignment is illustrated in equation \ref{formula:2}, where $T_y$ and $T_x$ represent the target and source text, respectively (e.g., [“face with smile”, “face”]). At this stage, we only need to ascertain the CLIP feature direction between the original and edited images. As the model continues to train, the generated image solely changes in this direction, ensuring that other local regions remain unaffected.

\begin{equation}
\mathcal{L}_{CLIP}^{direction}=1-\cos (E_i(y)-E_i(x), E_t(T_y)-E_t(T_x))
\label{formula:2}
\end{equation}

Furthermore, in the feature extraction stage of the input image, we introduce a fusion modulation module to integrate the text conditions into the input feature map of the hypernetworks. By modifying the numerical characteristics of the intermediate layer feature map $\bar{x}$, we achieve indirect control over hypernetworks, allowing it to generate various weight factors and enabling a single model to produce diverse editing effects. In the fusion modulation module, to maintain consistency between the text condition and the generated image, we also incorporate the CLIP feature direction $\Delta t$ between the texts as the conditional embedding. The embedding process is as follows:

\begin{equation}
\hat{x}=\bar{x} \cdot \alpha(\Delta t)+\beta(\Delta t), \text{where} \Delta t=E_t\left(T_y\right)-E_t\left(T_x\right)
\end{equation}

Here, $\alpha(\cdot)$ and $\beta(\cdot)$ refer to the multi-layer perceptrons (MLPs), and $\bar{x}$ denotes the feature map obtained from the input image $x$ through a series of operations such as convolution, batch normalization, and ResNet34 \cite{he2016deep}.

\begin{figure*}
    \centering
    \includegraphics[width=0.7\textwidth]{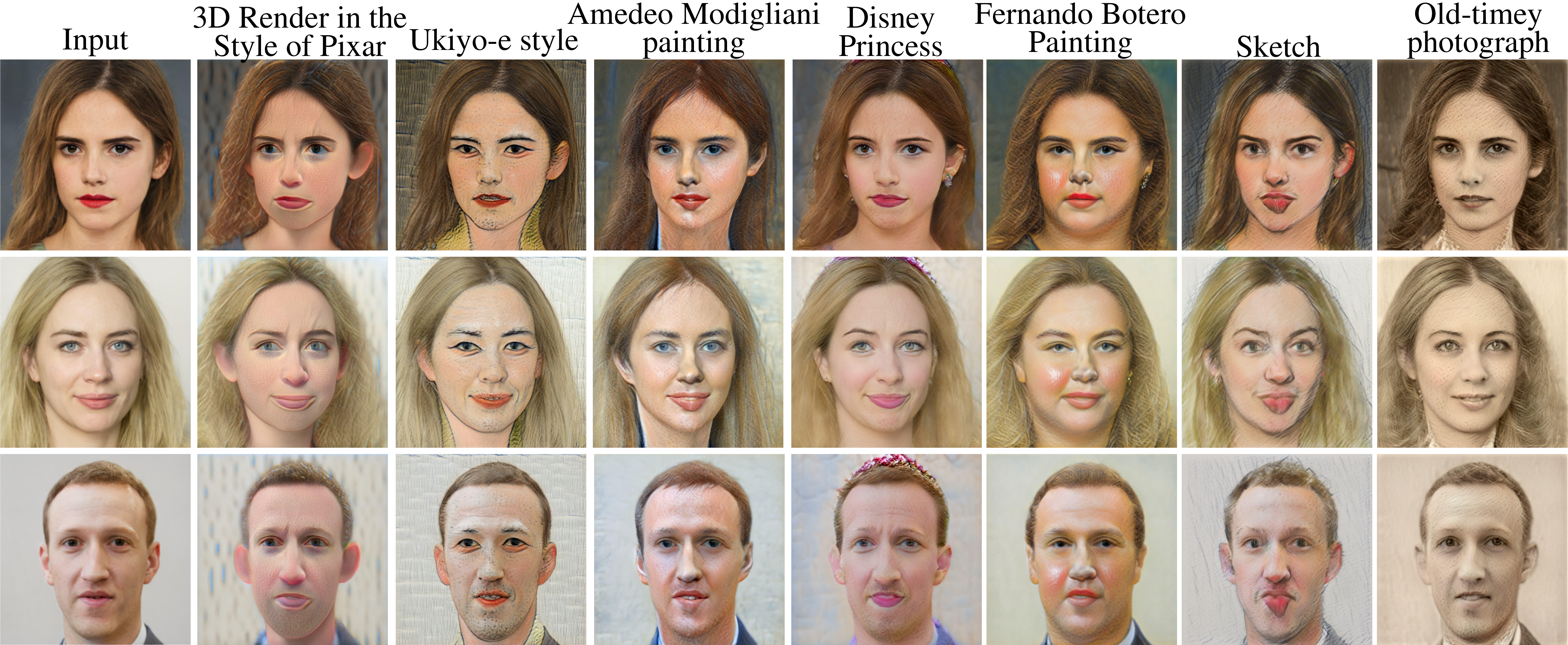}
    \caption{Results of cross-domain style editing by using our method. The target styles are located above the images.}
    \label{fig:edit_out_domain}
    
\end{figure*}

\subsection{Adaptive layer selector}
Recently, \cite{xia2021tedigan} demonstrated that different layers of the generator in StyleGAN2 control different attributes. Inspired by their findings, we question whether weight factors for all layers play a role in editing a single attribute. To explore this, we monitored how the weight factors generated by the hypernetworks changed during training. Figure \ref{fig:adapt-layers-choose-final} indicates that only a few layers undergo significant changes in the weight factors. Thus, we propose the adaptive layer selector, which allows hypernetworks to determine the layers requiring output weight factors autonomously. This technique alleviates the model's parameter count and maximizes the hypernetworks' efficiency when a single model only needs to implement a single attribute edit operation.

The schematic diagram of the adaptive layer selector is illustrated in Figure \ref{fig:adapt_layer_choose}. Before training the model, we identify layers that exhibit significant variations in latent codes through latent optimization \cite{patashnik2021styleclip}. These layers are then used as the selected layers to output weight factors. We found that the process of optimization can be completed with a few iterations, taking approximately less than 5 seconds. We first sample random noise \textit{$Z\sim N(0,1)$} to generate the initial latent codes $\omega_i^0$. Then we optimize the initial latent codes $\omega_i^0$ for $m$ iterations using the directional CLIP loss to obtain $\omega_i^m$. Now, we can obtain the difference set between the latent codes before and after optimization as $\Delta \omega_i=\left|\omega_i^m-\omega_i^0\right|$. To adaptively select the suitable layer, we establish an adaptive threshold as follows:

\begin{equation}
\varphi=\frac{\sum_{i=1}^n \Delta \omega_i}{n}+\lambda_{s t d} \sqrt{\frac{\sum_{j=1}^n\left(\Delta \omega_j-\frac{\sum_{i=1}^n \Delta \omega_i}{n}\right)^2}{n}}
\end{equation}

Where $\lambda_{std}$ represents the trade-off coefficient, and $n=17$. If $\Delta \omega_i \geq \varphi$, then the $i^{th}$ layer is considered the layer requiring output weight factors, and its index is added to the sequence $L$. Otherwise, no further action is taken. Finally, we obtain the sequence $L=\{i_1, i_2, \ldots, i_n\}$.

\begin{figure*}
    \centering
    \includegraphics[width=\textwidth]{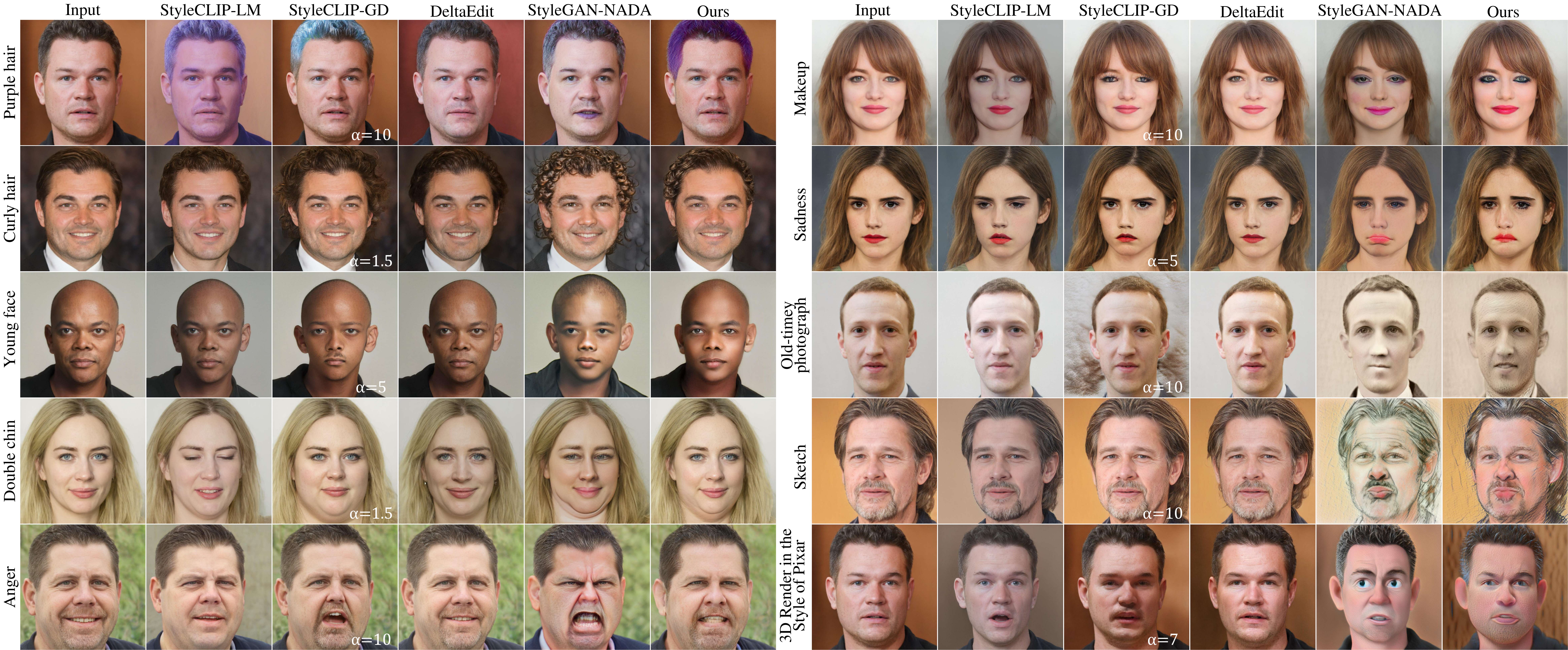}
    \caption{The results of comparing our method with DeltaEdit \cite{lyu2023deltaedit}, StyleCLIP-LM \cite{patashnik2021styleclip}, StyleCLIP-GD \cite{patashnik2021styleclip}, and StyleGAN-NADA \cite{gal2022stylegan} for image editing.}
    \label{fig:comparable_cele_hq}
    
\end{figure*}

\subsection{Loss functions}
Our objective is to modify specific target regions of the images while preserving the non-target regions unchanged. To achieve this, we follow the previous approach \cite{alaluf2022hyperstyle} and use a similarity loss in our work. The calculation process is as follows:

\begin{equation}
\begin{split}
\mathcal{L}_{Sim}=1- {cos} ( &R_{Sim}(G(\theta+\Delta \cdot \theta , w_{init})), \\ &R_{Sim}(G(\theta, w_{init})))
\end{split}
\label{equation:7}
\end{equation}

$R_{Sim}$ represents the pre-trained ArcFace network \cite{deng2019arcface} when the initial images belong to the facial domain, and the pre-trained MoCo model \cite{he2020momentum} when initial images belong to the non-facial domain. Additionally, we minimize the variations in global regions through L2 loss. The calculation process is as follows:

\begin{equation}
\mathcal{L}_2=\left\|G\left(\theta+\Delta \cdot \theta, w_{init}\right)-G\left(\theta, w_{init }\right)\right\|_2
\end{equation}

Combined with our goal of text-driven image editing, we define our comprehensive loss function as follows:

\begin{equation}
\mathcal{L}=\lambda_{clip} \mathcal{L}_{CLIP}^{direction}+\lambda_{norm} \mathcal{L}_2+\lambda_{Sim} \mathcal{L}_{Sim}
\end{equation}

Where $\lambda_{clip}$ and $\lambda_{norm}$ are both set to 1. And $\lambda_{Sim}$ can take the values of either 0.1 or 0.5, depending on whether $R_{Sim}$ corresponds to the ArcFace or MoCo networks. Notably, the ArcFace and MoCo networks cannot be simultaneously used during the model training process.

\section{Experiments}
\subsection{Implementation details}
To validate the effectiveness of our approach, we conducted extensive experiments on diverse and challenging datasets. For the face domain, we utilized the FFHQ \cite{Karras2018ASG} dataset with 70,000 images as the training set and the Celeba-HQ \cite{karras2018progressive} dataset with 28,000 images as the test set. Additionally, in the supplementary material, we provided image editing results on the Cat, Horse, Car, and Church datasets of LSUN \cite{yu2015lsun}, as well as the Cat, Dog, and Wild datasets of AFHQ \cite{choi2020stargan}. It is worth noting that all real images were inverted using the e4e encoder \cite{tov2021designing} to obtain the latent codes, and all generated images were produced using the pre-trained StyleGAN2 generator. Our training was conducted on a 4090 GPU, with a batch size of 4, and the Ranger optimizer, using a learning rate of 0.001.

\subsection{Qualitative evaluation}
\subsubsection{Results of authentic images editing.}
To validate the effectiveness of our method in utilizing various text conditions for image editing, we present the results of using text to control 18 different attributes in the image, as shown in Figure \ref{fig:ffhq_edit}. The results demonstrate that our method effectively controls specific attributes while preserving irrelevant ones. More editing results from various datasets are shown in the supplementary material.

\subsubsection{Results of cross-domain images editing.}
Figure \ref{fig:edit_out_domain} shows the results of our method editing real images to target images of different domains. The target domain never appears in the training process, which indicates that our method has good domain generalization ability. More editing results of cross-domain images editing are shown in the supplementary material.

\begin{figure}
    \centering
    \includegraphics[width=0.85\linewidth]{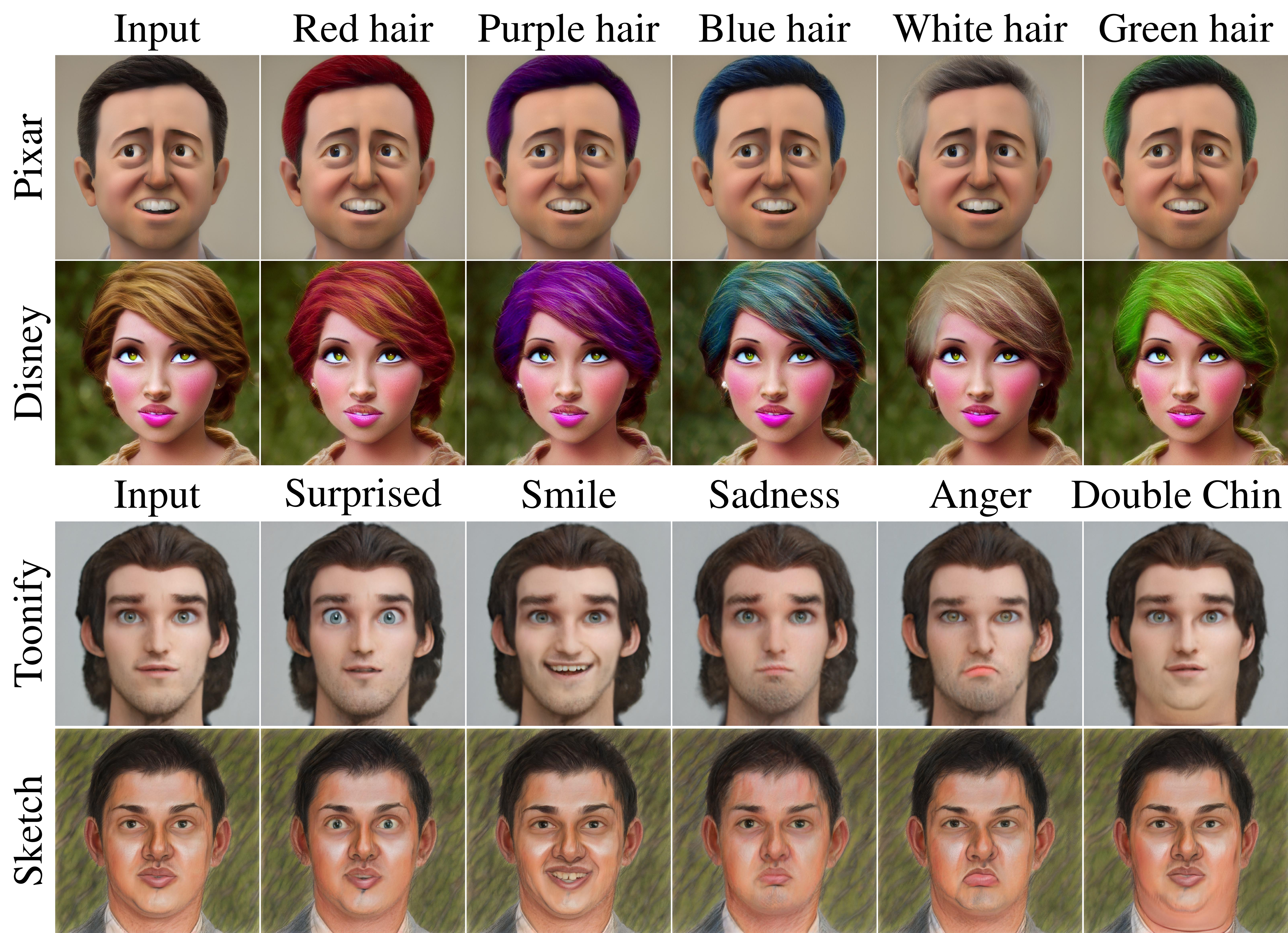}
    \caption{The weight factors produced by HyperEditor trained on the FFHQ dataset are applied to other domain generators (e.g., StyleGAN-NADA \cite{gal2022stylegan}).}
    \label{fig:adapt_edit}
\end{figure}

\subsubsection{Comparisons with the SOTA.}
In Figure \ref{fig:comparable_cele_hq}, We compare our method with three current state-of-the-art approaches, namely DeltaEdit \cite{lyu2023deltaedit}, StyleGAN-NADA \cite{gal2022stylegan}, and StyleCLIP \cite{patashnik2021styleclip}. Among these methods, DeltaEdit and StyleCLIP utilize the manipulation of latent codes for image editing. In contrast, StyleGAN-NADA achieves style transfer by fine-tuning the generator's weights through retraining. In StyleCLIP, due to the extensive time consumption caused by optimization-based methods, here we only consider two methods based on latent mapper and global directions, respectively named StyleCLIP-LM and StyleCLIP-GD. Compared with DeltaEdit, where the editing results remain unchanged from the input image for attributes like “Purple hair” and “Young face”, our method excels at achieving accurate modifications of specific image attributes. Compared with StyleCLIP-LM, our method not only edits more accurately, but also protects non-relevant regions better. Compared with StyleCLIP-GD, our method achieves better image editing results without parameter adjustment. Furthermore, compared to the approaches that manipulate the latent codes, our method can accomplish cross-domain image editing, while they cannot. Nevertheless, while StyleGAN-NADA excels in style transfer, our method outperforms it regarding the controllability of detailed attribute editing and the preservation of facial identity. More qualitative comparison results are shown in the supplementary material.

\subsubsection{Weight factors' transferability.}
In this section, we apply the weight factors trained on the FFHQ dataset to generators in various domains. The edited images are depicted in Figure \ref{fig:adapt_edit}. The results demonstrate that the generated weight factors can be effectively transferred to generators in other domains, enabling authentic facial attribute editing without compromising the target style. More results are shown in the supplementary material.

\begin{table*}[htbp]
  \centering
  
    \begin{tabular}{lp{1cm}<{\centering}p{1cm}<{\centering}p{1cm}<{\centering}p{1cm}<{\centering}p{1cm}<{\centering}p{1cm}<{\centering}cc}
    \toprule
    Methods & \multicolumn{1}{c}{PSNR↑} & \multicolumn{1}{c}{LPIPS↓} & \multicolumn{1}{c}{SSIM↑} & \multicolumn{1}{c}{IDS↑} & \multicolumn{1}{c}{FID-0↓} & \multicolumn{1}{c}{FID-1↓} & \multicolumn{1}{c}{CS↑} & \multicolumn{1}{c}{Nop↓} \\
    \midrule
    StyleCLIP-GD $\alpha=5 $ & 24.96 & 0.28  & 0.79  & 0.72  & 11.73 & 44.05 & 24.82 & -\\
    StyleCLIP-GD $\alpha=10 $ & 20.51 & 0.33  & 0.76  & 0.64  & 17.08 & 201.95 & 25.28 & -\\
    StyleCLIP-LM & 20.57 & 0.23  & 0.81  & 0.84  & 8.49 & 14.58   & 26.35 & 33.52M\\
    StyleGAN-NADA  & 18.75 & 0.27  & 0.74  & 0.58  & 56.20 & 59.54   & 26.69 & -\\
    DeltaEdit & 23.31 & 0.23  & 0.82  & 0.81  & 10.04 & 8.6   & 22.89 & 82.76M\\
    \midrule
    Ours  & \textbf{25.33} & \textbf{0.22} & \textbf{0.85} & \textbf{0.85} & \textbf{6.19} & \textbf{7.19} & \textbf{27.35} & 71.48M\\
    Ours(Global-CLIP) & 23.21 & 0.39  & 0.77  & 0.71  & 120.18 & 61.07 & 22.96 & -\\
    Ours(w/ adaptive layer selector) & 24.87 & 0.18 & 0.87 & 0.83  & 8.84  & 7.5   & 26.37 & 15.66M \\
    \bottomrule
    \end{tabular}%
    \caption{Quantitative evaluation of edited face images. Where FID-0 and FID-1 are obtained by calculating 2000 images before and after editing the “smile” and “double chin” attributes, respectively, the other values are obtained by computing the mean of images before and after editing for the ten different facial attributes. Nop denotes the number of model parameters.}
    \label{tab:addlabel}%
  
\end{table*}%

\subsection{Quantitative evaluation}
In Table \ref{tab:addlabel}, we provide the objective measures, including FID, PSNR, SSIM, LPIPS, IDS (identity similarity), and CS (CLIP score). All the results are the average values obtained by calculating the images before and after changing the ten attributes. Compared with state-of-the-art methods, our approach achieves the highest CLIP score (\textbf{27.35}), indicating that our results are more consistent with the target condition, confirming that HyperEditor can conduct more authentic image editing operations. Furthermore, in addition to achieving authentic editing, our method excels at preserving the image regions that are irrelevant to the editing target (as evidenced by Table \ref{tab:addlabel}, where we achieve the best results in the first four columns). Moreover, we calculated the FID values for the variations of “smile” and “double chin” attributes, which are displayed in Table \ref{tab:addlabel}. The minimal FID values signify the closest distribution between the images produced by our method and the original images, and it also reflects the protection of non-edited regions by our approach.

\begin{figure}
    \centering
    \includegraphics[width=\linewidth]{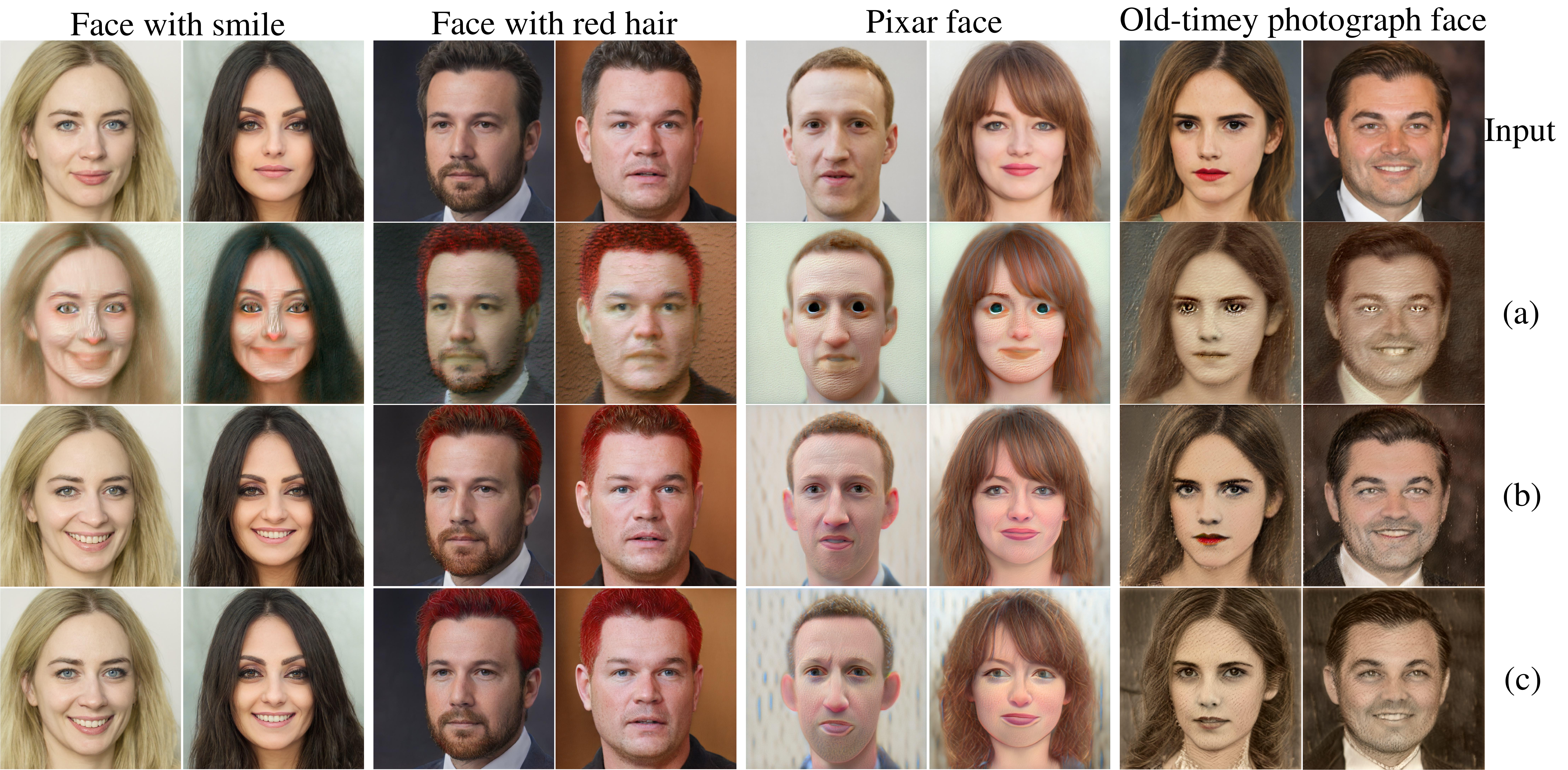}
    \caption{The results are presented in three cases: (a) without the adaptive layer selector and with Global-CLIP guidance, (b) with the adaptive layer selector and Directional-CLIP guidance, and (c) without the adaptive layer selector and with Directional-CLIP guidance.}
    \label{fig:ablation}
\end{figure}

\subsection{Ablation studies}
\subsubsection{Effectiveness of directional CLIP loss.}
The Global-CLIP guided text-driven image editing results are presented in Table \ref{tab:addlabel} and Figure \ref{fig:ablation}. The results indicate that the Global-CLIP guided method disrupts the global characteristics of the original image, leading to either significant differences between the generated image and the original image or blurriness. We attribute this to CLIP causing perturbations to the global feature semantics while guiding the local feature semantics to change. Additionally, \cite{gal2022stylegan} mentions the occurrence of adversarial interference during the image generation process guided by Global-CLIP. In contrast, our directional CLIP loss effectively protects the global feature semantics and provides more stable supervised training.

\begin{figure}
    \centering
    \includegraphics[width=\linewidth]{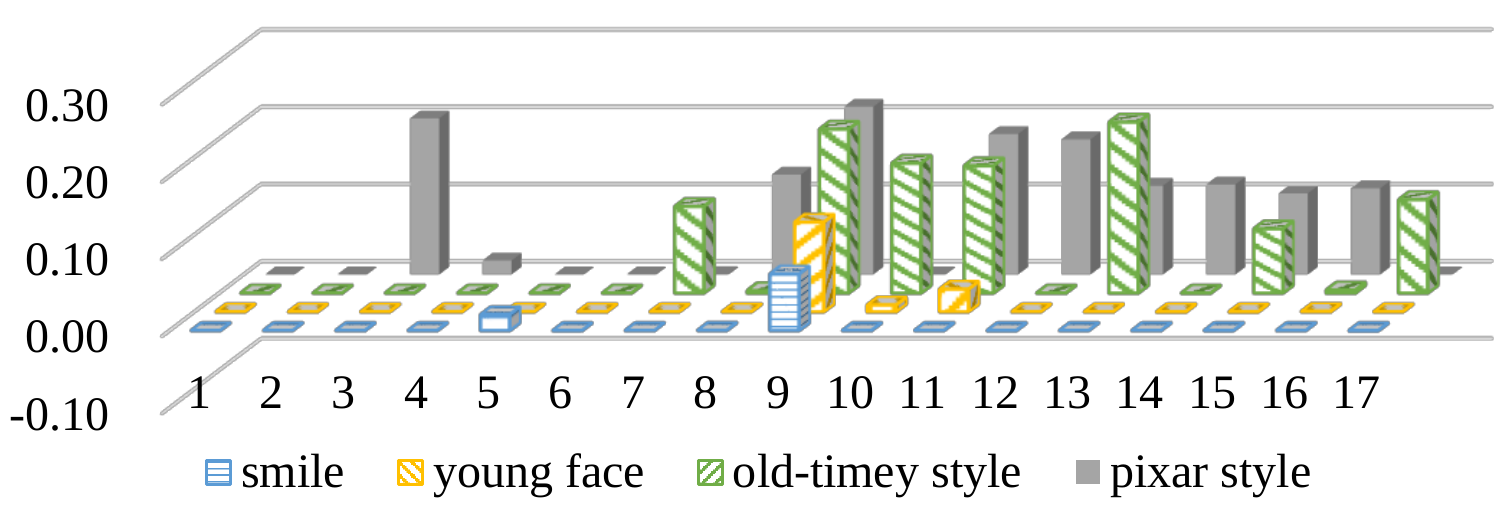}
    \caption{The variation in the average weight factors for each layer before and after editing specific attributes. The horizontal axis represents the layer index.}
    \label{fig:adapt-layers-choose-final}
    
\end{figure}

\subsubsection{Rationality of adaptive layer selector.}
In Figure \ref{fig:adapt-layers-choose-final}, we have documented the fluctuations in the average weight factors for each layer from the initiation of hypernetwork training until convergence. The results reveal that only some layers' weight factors undergo changes, confirming that it is not necessary to output weight factors for all layers during the image editing process. Table \ref{tab:addlabel} and Figure \ref{fig:ablation} present results of image editing when we utilize the adaptive layer selector, at this point, $\lambda_{std}=0.6$. The results demonstrate that using the adaptive layer selector can reduce the parameter count of hypernetworks by $80\%$, while still achieving equivalent results. Note that the adaptive layer selector is only suitable for a single model to edit a single attribute. If a single model is used to edit multiple attributes, the weight factors of all layers can be effectively used, so there is no need to reduce the number of layers that output weight factors. The selection of $\lambda_{s t d}$ is illustrated in the supplementary material.

\section{Conclusions}
In this paper, we propose HyperEditor, an innovative framework that leverages hypernetworks to reassign weights of StyleGAN2's generator and utilizes CLIP for supervised training. As a result, compared with previous methods that achieve image editing by manipulating latent codes, our HyperEditor enables both authentic attribute editing and cross-domain style transfer. Compared to fine-tuning the generator by retraining, reassigning the generator's weights using hypernetworks offers more excellent controllability, enabling it to achieve finer precision in attribute editing and safeguarding the coherence of non-edited regions. Moreover, our fusion modulation module allows diverse editing operations within a single model, and the adaptive layer selector can reduce the model's complexity while editing a single attribute. Our innovative approach will open up the possibility to edit one thing to anything in the future.

\section{Acknowledgments}
This work was supported by the National Natural Science Foundation of China under Grant 61871186 and 61771322.

\section{Appendix Materials}
\subsection{Results of images editing by our method}
In this section, we present comprehensive qualitative results that further demonstrate the efficacy, versatility, and applicability of our HyperEditor.

\subsubsection{Results of cross-domain image editing}
\begin{figure*}
    \centering
    \includegraphics[width=\textwidth]{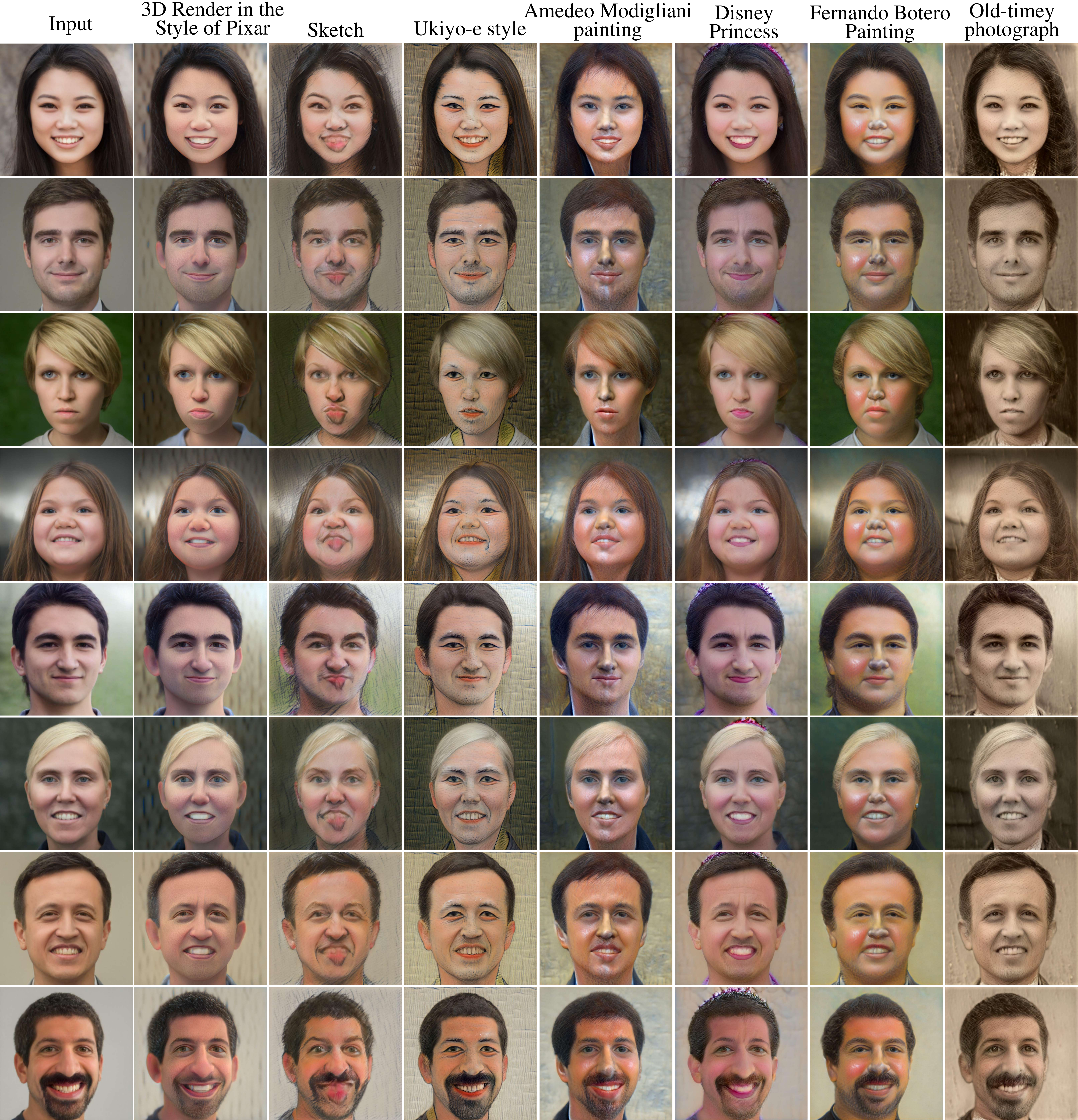}
    \caption{We use our method to edit the images of the FFHQ dataset to various out-of-domain images. The text prompt containing the target style is located above each column.}
    \label{fig:suplement_outdomain}
\end{figure*}

In Figure \ref{fig:suplement_outdomain}, we showcase additional visual results of image editing across various domains, achieved through our approach on the FFHQ dataset. During the training process, no target domain images are present, proving our method's excellent generalization.

\subsubsection{More visual results on different datasets}
\begin{figure*}
    \centering
    \includegraphics[width=\textwidth]{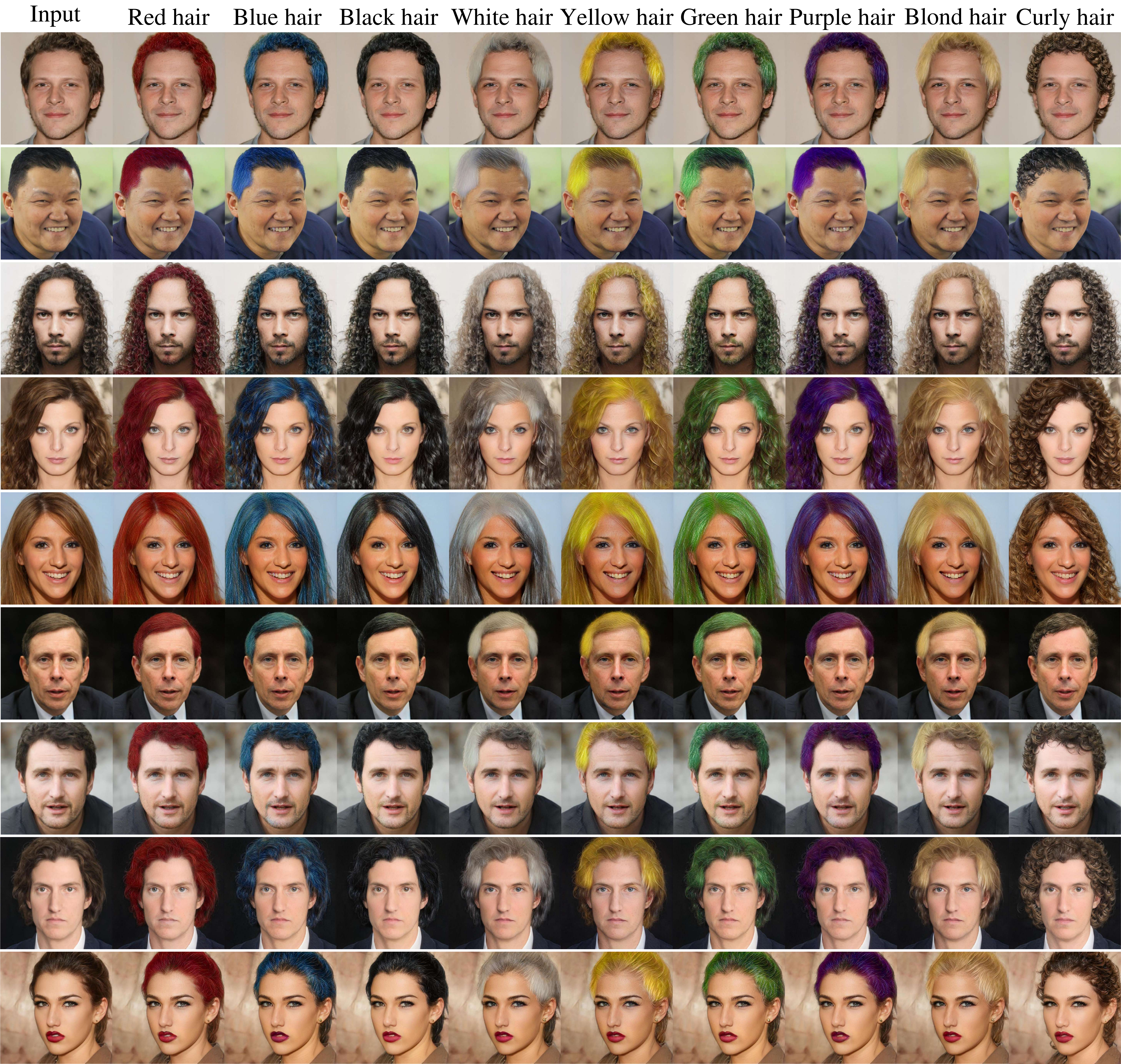}
    \caption{Various hair colors and hairstyles on the Celeba-HQ dataset were edited using our method. The text prompt containing the target attribute is located above each column.}
    \label{fig:suplement_edit}
\end{figure*}

\begin{figure*}
    \centering
    \includegraphics[width=0.9\textwidth]{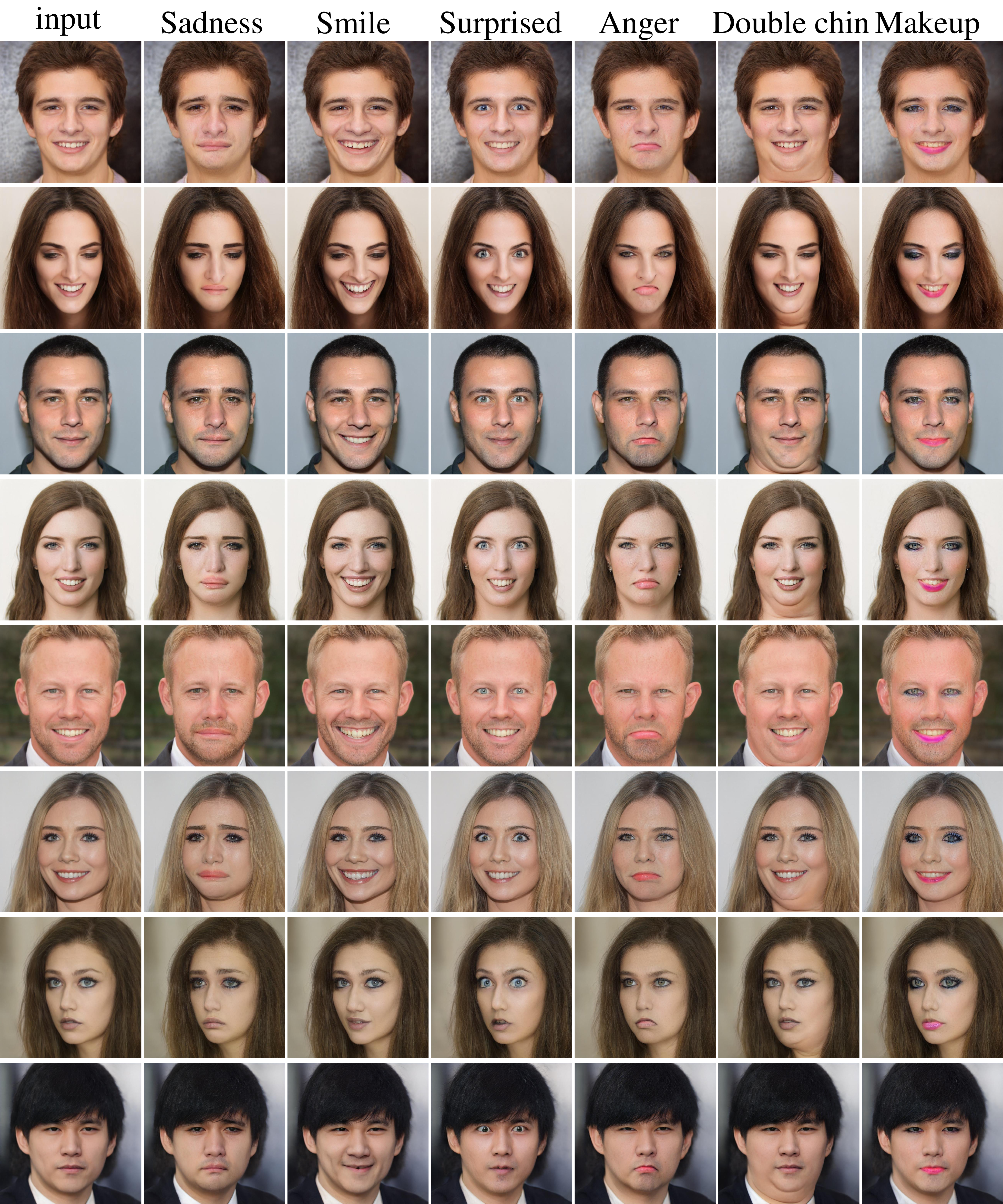}
    \caption{The facial images of the Celeba-HQ dataset are edited in various ways using our method. The text prompt containing the target attribute is located above each column.}
    \label{fig:suplement_edit1}
\end{figure*}

\begin{figure*}
    \centering
    \includegraphics[width=0.9\textwidth]{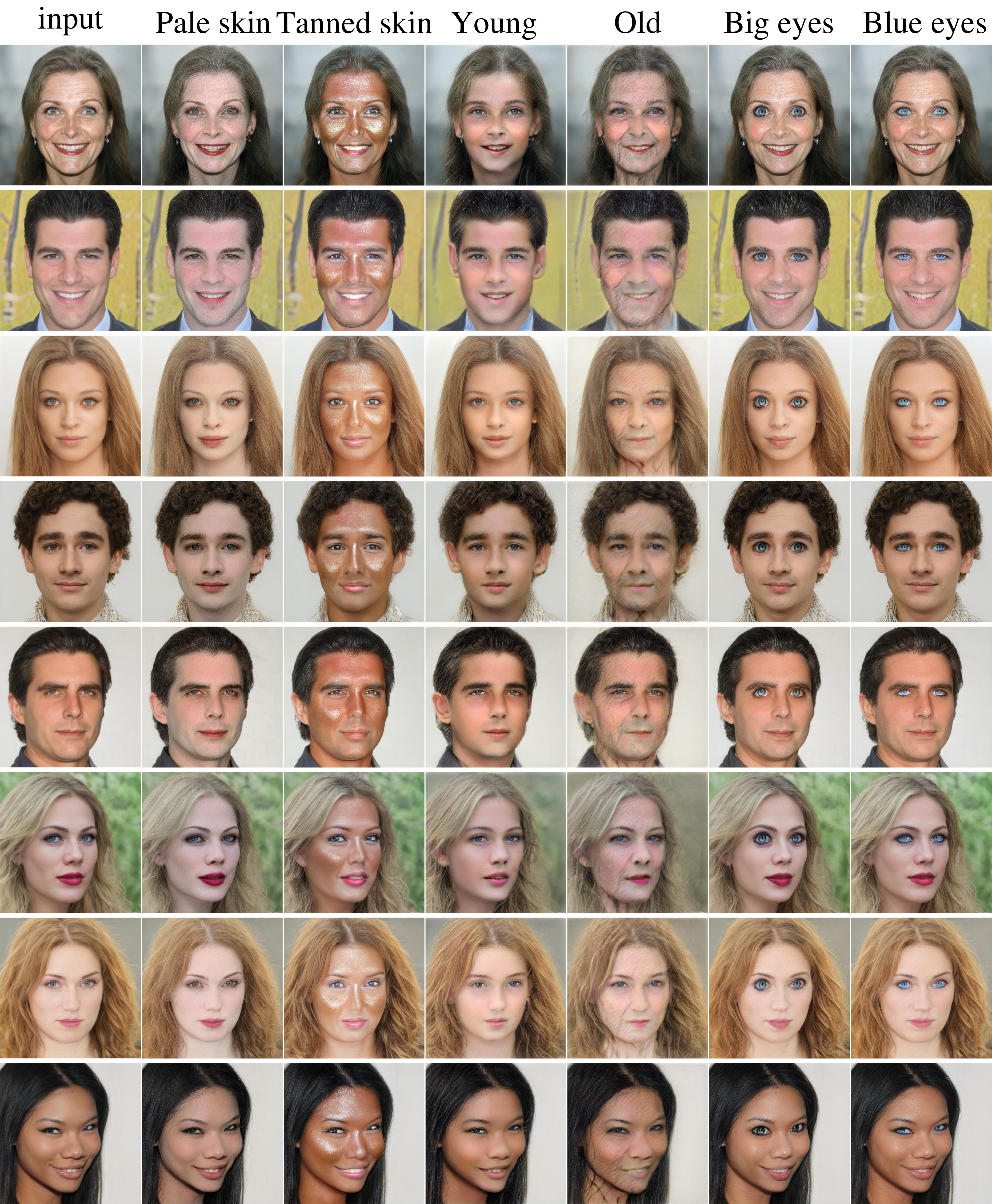}
    \caption{The facial images of the Celeba-HQ dataset are edited in various ways using our method. The text prompt containing the target attribute is located above each column.}
    \label{fig:suplement_edit2}
\end{figure*}

\begin{figure*}
    \centering
    \includegraphics[width=0.9\textwidth]{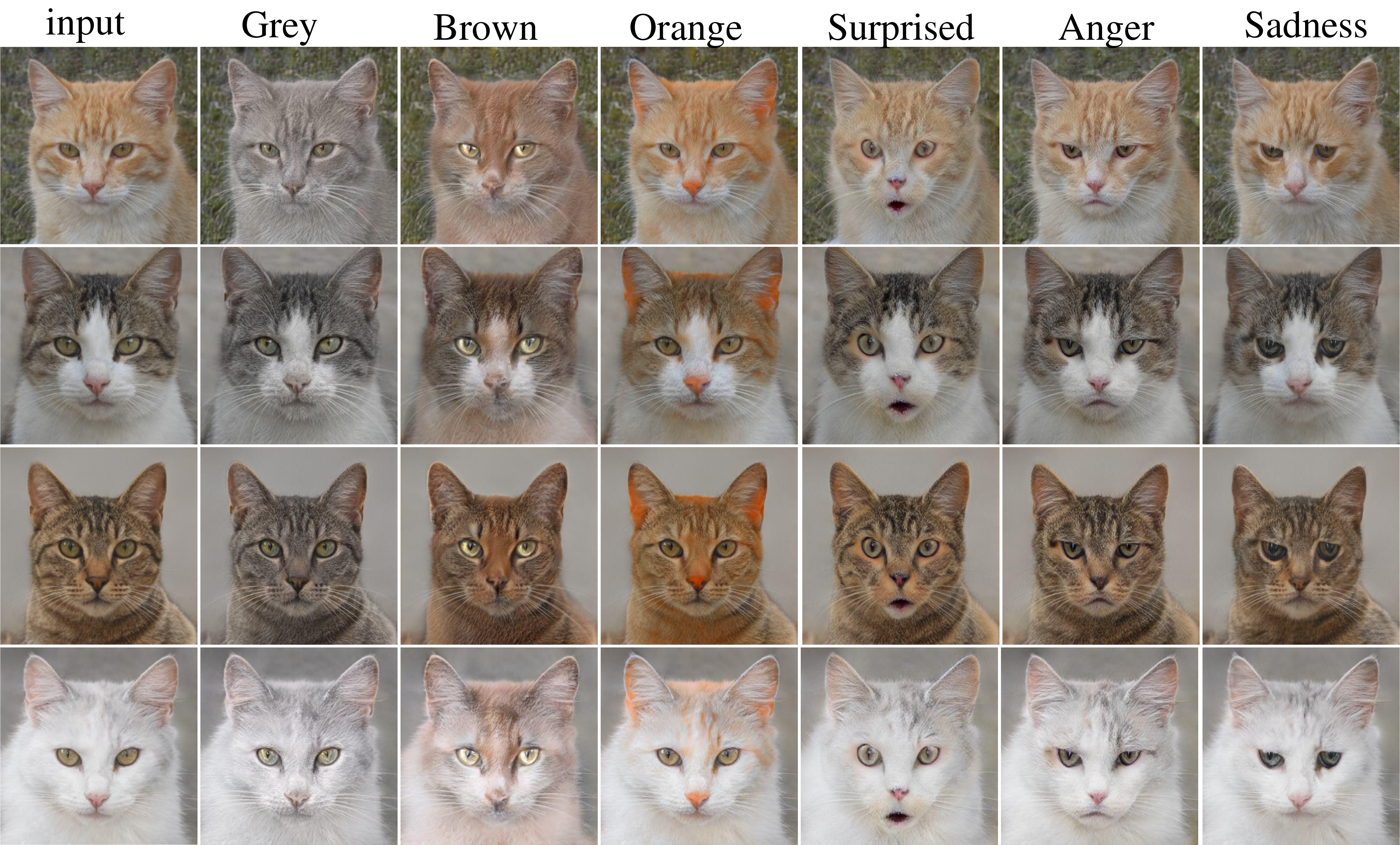}
    \caption{Diverse image editing using our method on the AFHQ-cat dataset. The text prompt containing the target attribute is located above each column.}
    \label{fig:cat}
\end{figure*}

\begin{figure*}
    \centering
    \includegraphics[width=0.9\textwidth]{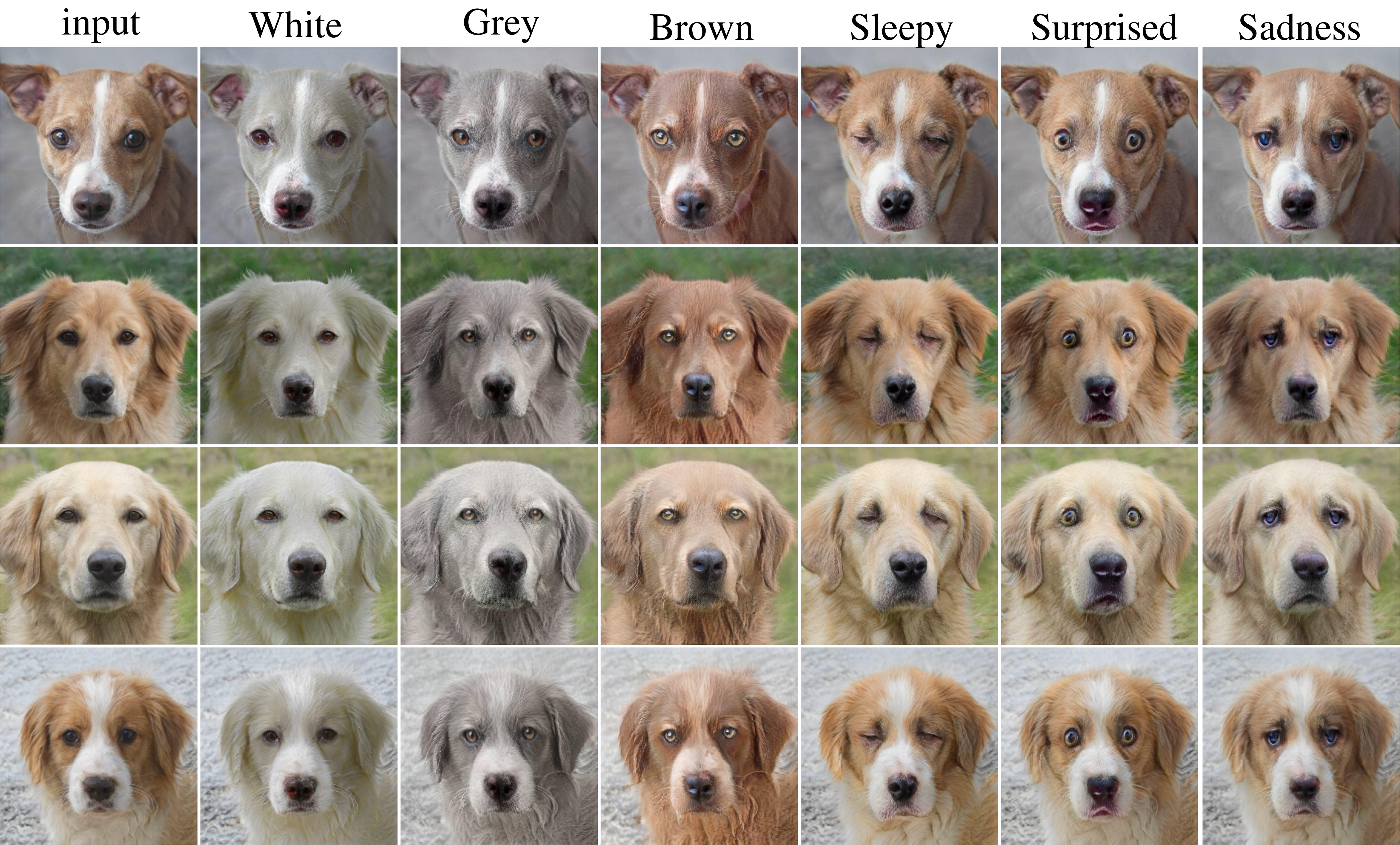}
    \caption{Diverse image editing using our method on the AFHQ-dog dataset. The text prompt containing the target attribute is located above each column.}
    \label{fig:dog}
\end{figure*}

\begin{figure*}
    \centering
    \includegraphics[width=0.9\textwidth]{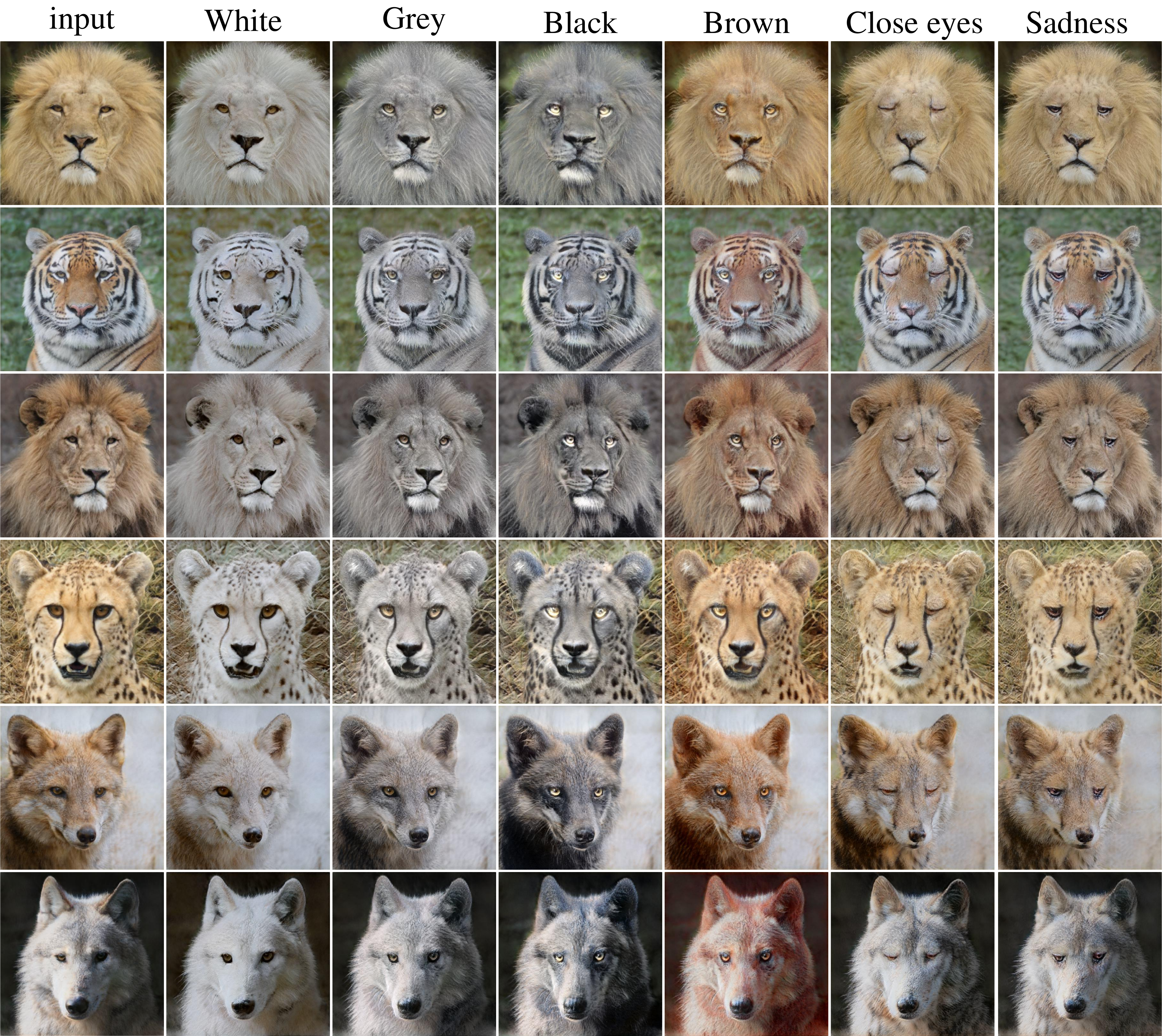}
    \caption{Diverse image editing using our method on the AFHQ-wild dataset. The text prompt containing the target attribute is located above each column.}
    \label{fig:wild}
\end{figure*}

\begin{figure*}
    \centering
    \includegraphics[width=\textwidth]{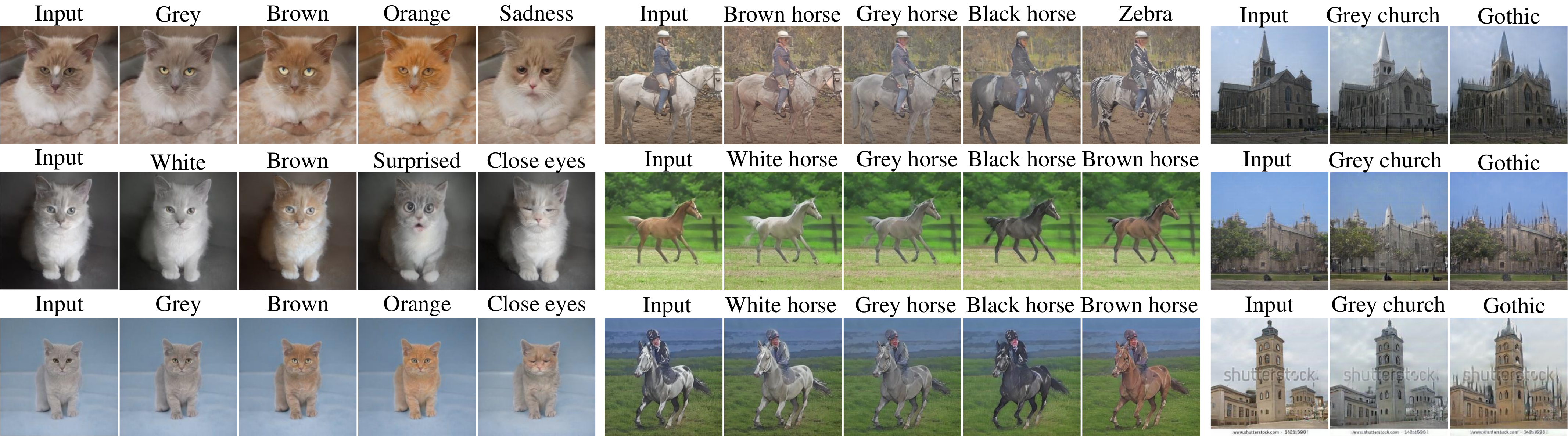}
    \caption{Various image edits were performed on LSUN cat, horse, and church datasets using our method. The target attribute specified in the text prompt is shown above each image.}
    \label{fig:lsun}
\end{figure*}

\begin{figure*}
    \centering
    \includegraphics[width=\textwidth]{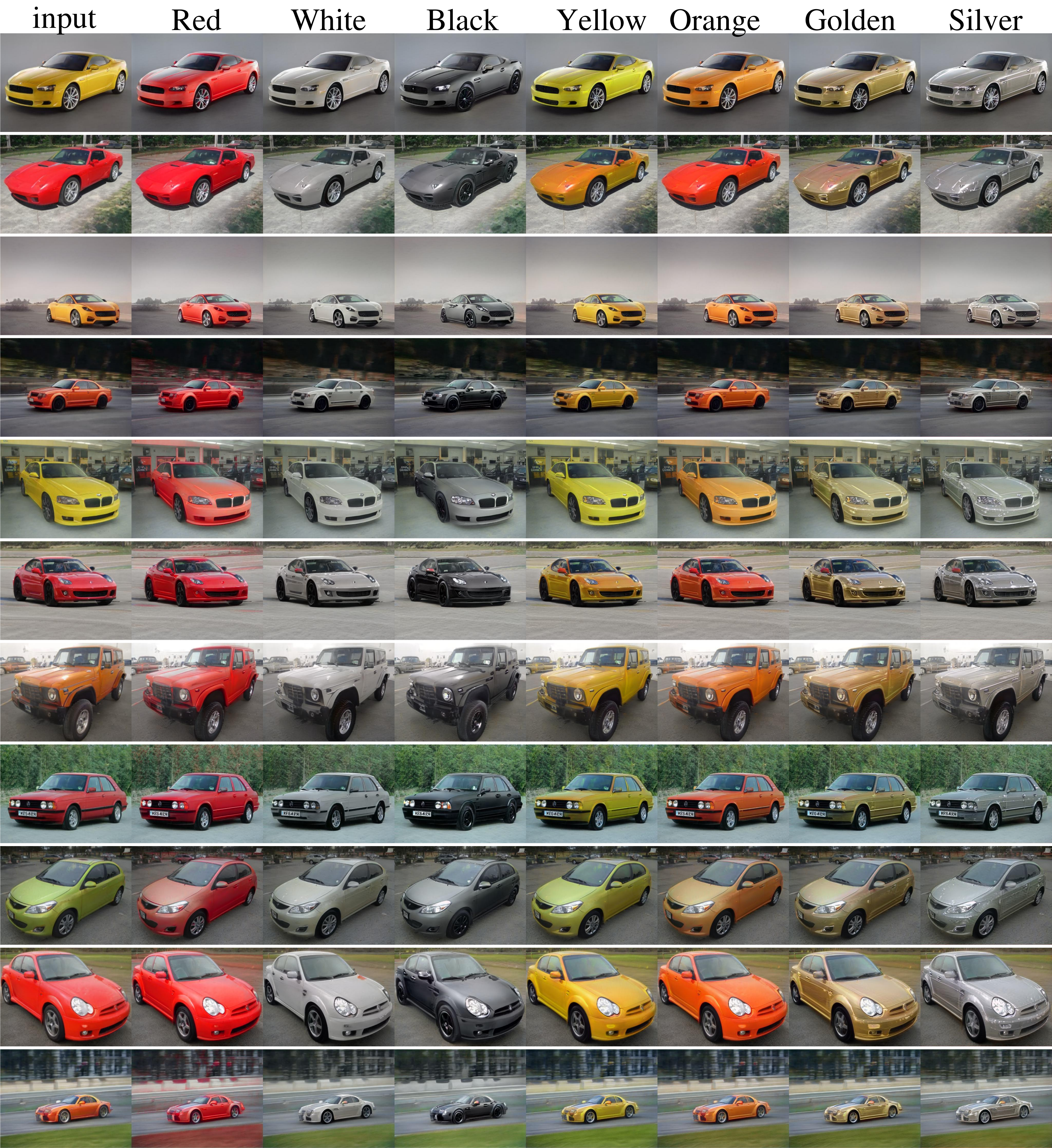}
    \caption{Using our method to manipulate different colors on the LSUN-car dataset images. The text prompt contains the target attribute located above the image.}
    \label{fig:car}
\end{figure*}

In Figure \ref{fig:suplement_edit}, Figure \ref{fig:suplement_edit1} and Figure \ref{fig:suplement_edit2}, we show more visual results of image editing using our approach on the Celeba-HQ dataset. Furthermore, we present more visual results from the AFHQ cat, dog, and wild datasets in Figure \ref{fig:cat}, Figure \ref{fig:dog}, and Figure \ref{fig:wild}, respectively. Additionally, we include extensive visual results of image editing on the LSUN cat, horse, church, and car datasets, as shown in Figure \ref{fig:lsun} and Figure \ref{fig:car}. Our method exhibits authentic attribute editing capabilities across diverse datasets, demonstrating its versatility and effectiveness.

\subsubsection{Simultaneous multi-attribute image editing}
\begin{figure*}
    \centering
    \includegraphics[width=\textwidth]{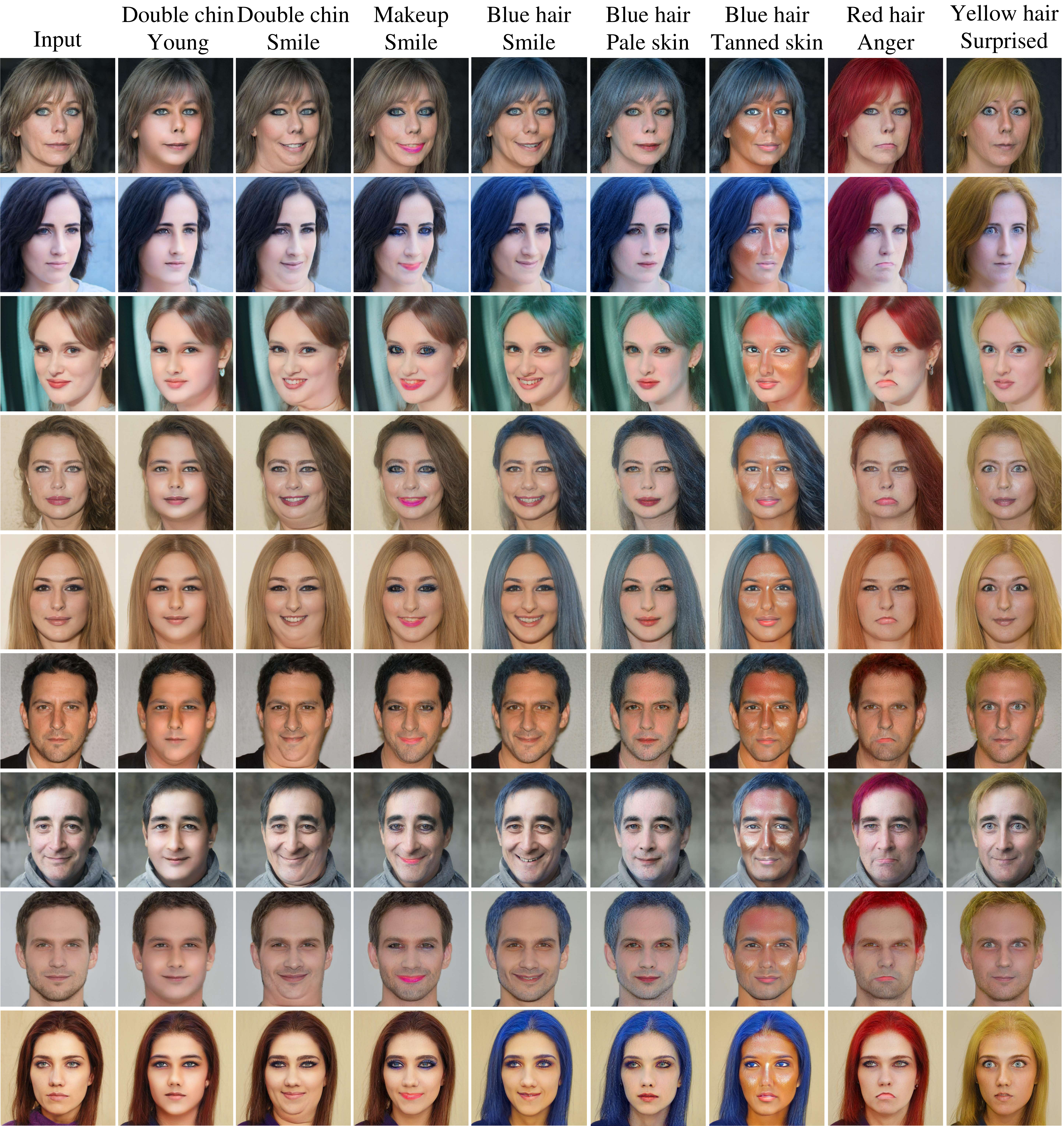}
    \caption{The result of editing two attributes on an image. The target attribute specified in the text prompt is shown above each image.}
    \label{fig:mix_edit}
\end{figure*}

We present the visual results of simultaneously editing two attributes in Figure \ref{fig:mix_edit}. This showcases our method's ability to modify only the target attribute regions during image editing while ensuring that unrelated regions remain unchanged.

\subsubsection{More visual results of weight factors transfer}
\begin{figure*}
    \centering
    \includegraphics[width=\textwidth]{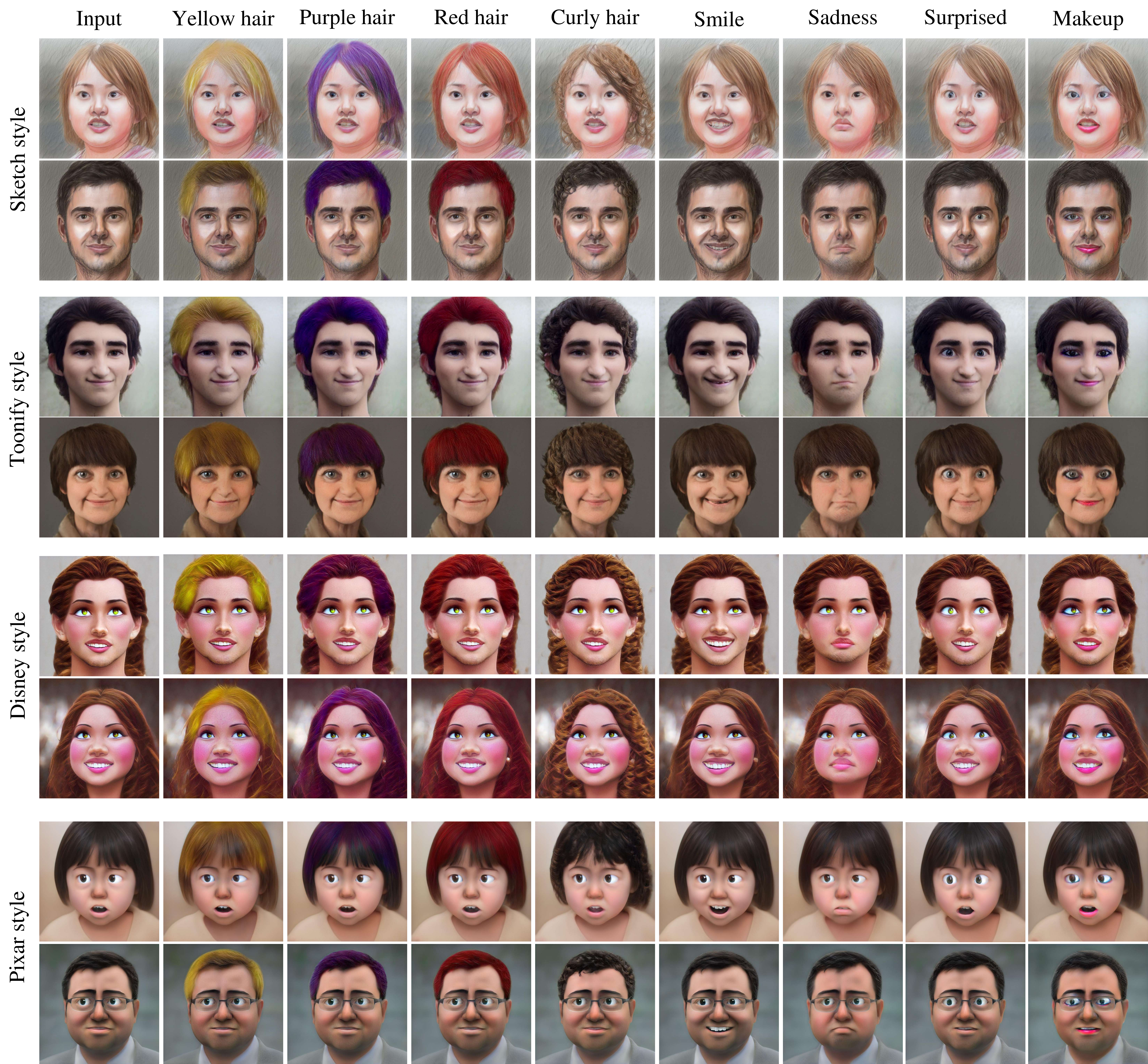}
    \caption{Image editing results by transferring weighting factors trained on FFHQ to generators in other image domains. The name of the image domain is located to the left of the image, while the target attribute is displayed above the image.}
    \label{fig:suplement_adapt_edit}
\end{figure*}

We transfer the weight factors trained on the FFHQ dataset to generators for other image domains and showcase the visual results of image editing for those domains in Figure \ref{fig:suplement_adapt_edit}. The precise image editing outcomes demonstrate the transferability of our approach.

\subsubsection{Attributes interpolation}
\begin{figure*}
    \centering
    \includegraphics[width=0.9\textwidth]{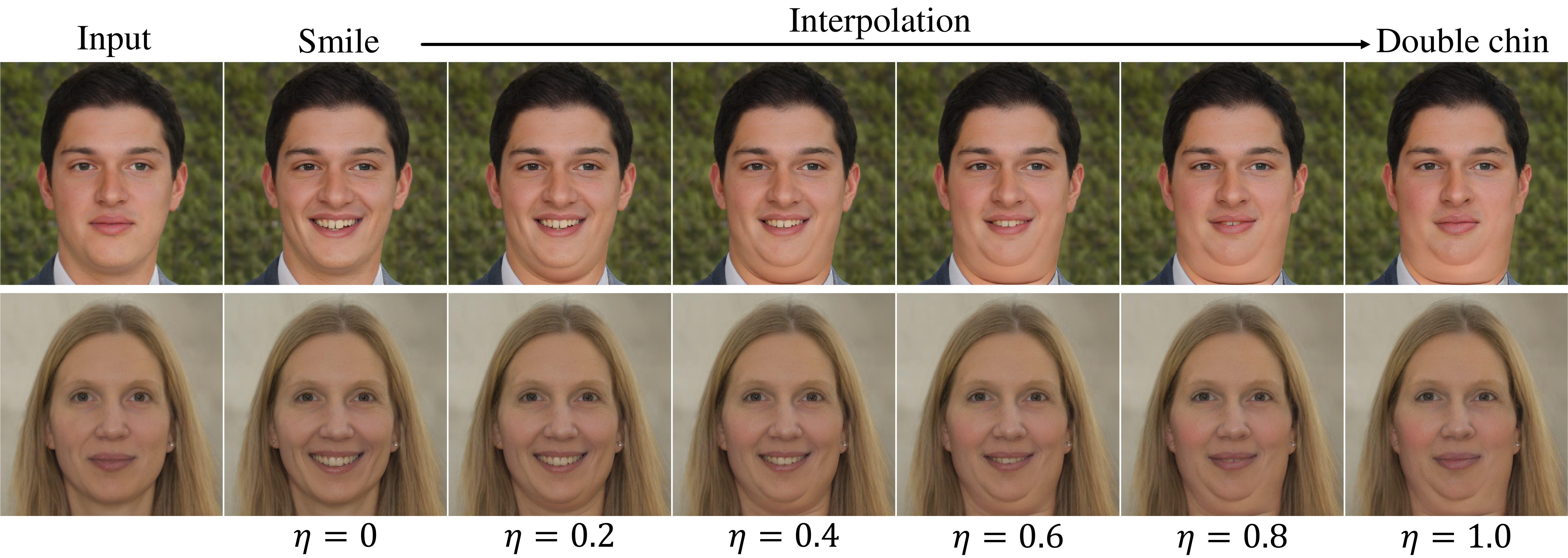}
    \caption{Results of facial attribute interpolation. By increasing $\eta$ from 0 to 1 in increments of 0.2, the style of the interpolated image is continuously transformed from attribute A (smile) to attribute B (Double chin).}
    \label{fig:intepolate1}
\end{figure*}

\begin{figure*}
    \centering
    \includegraphics[width=0.9\textwidth]{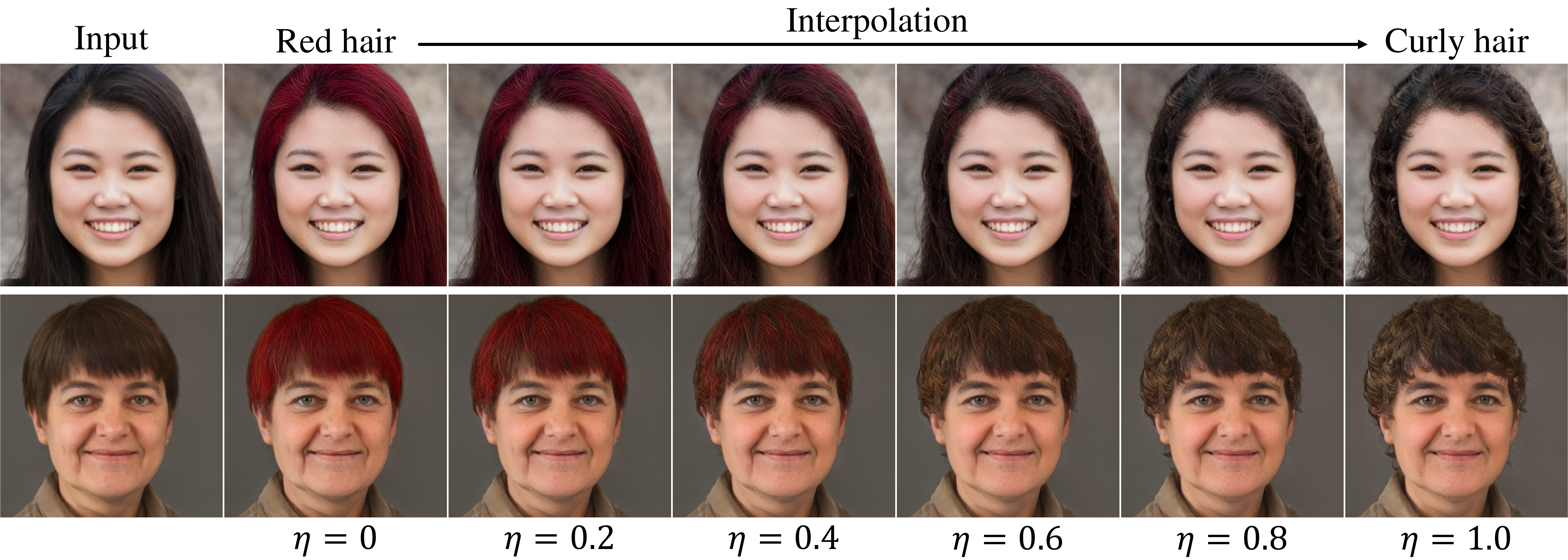}
    \caption{Results of hair attribute interpolation. The style of the interpolated image is continuously transferred from attribute A (Red hair) to attribute B (Curly hair) by incrementing $\eta$ from 0 to 1 at intervals of 0.2.}
    \label{fig:Interpolation}
\end{figure*}

Our approach exhibits comparable control interpolation capabilities to previous methods \cite{lyu2023deltaedit, wei2022hairclip}, which manipulate latent codes for image editing. By providing two distinct textual descriptions, our hypernetworks can generate two different weight factors, denoted as $\Delta_a^\prime$ and $\Delta_b^\prime$, enabling facial attribute manipulation through interpolation. The new weighting factor, $\Delta$, can be computed as $\Delta=\eta \cdot {\Delta_a^\prime}+(1-\eta)\cdot {\Delta_b^\prime}$, where $\eta$ serves as the interpolation parameter. As demonstrated in Figure \ref{fig:intepolate1} and Figure \ref{fig:Interpolation}, varying the parameter $\eta$ from 0 to 1 at intervals of 0.2 yields natural and impressive interpolation results between conditional text A (“Smile”) to conditional text B (“Double chin”) and conditional text A (“Red hair”) to conditional text B (“Curly hair”), respectively.

\subsubsection{How text prompts affect results}
\begin{figure*}
    \centering
    \includegraphics[width=0.9\textwidth]{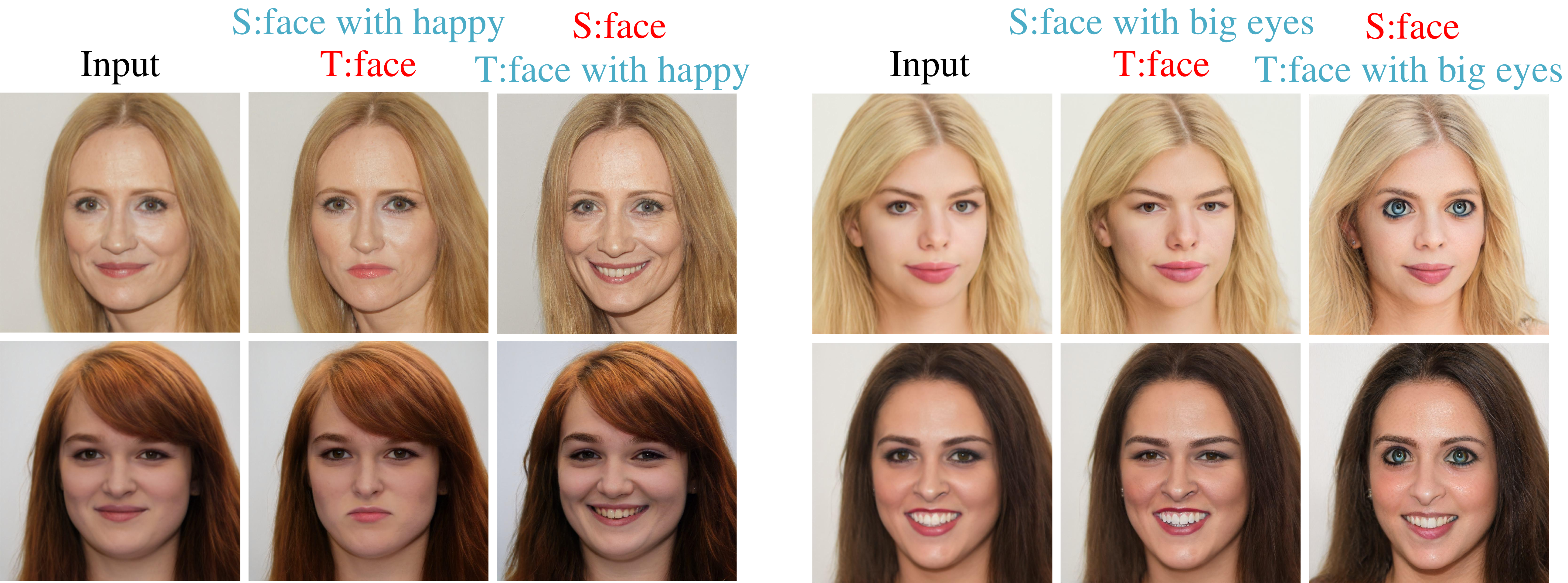}
    \caption{The image results are generated by the interchange of the target and source text prompt. The source prompt is in blue, while the target prompt is in red. When the target text prompt changes to the original text prompt, the generated attribute will be opposite to the target attribute.}
    \label{fig:condition}
\end{figure*}

As we employ text pairs, such as “face” and “face with smile”, as text conditional embeddings, we perform experiments to examine the image editing effect when interchanging the target text with the source text, as depicted in Figure \ref{fig:condition}. When the target text description is used as the source text, the results of image editing exhibit the opposite effect. For instance, if “face with smile” is used as the source text, the edited result obtained with this condition will display the state of “sadness”. It shows that our results produce variations based on the CLIP feature orientation.

\subsection{Compared with the SOTA}
In this section, we conduct additional experiments to compare our method with the DeltaEdit \cite{lyu2023deltaedit}, styleGAN-nada \cite{gal2022stylegan}, styleCLIP-GD \cite{patashnik2021styleclip}, and StyleCLIP-LM \cite{patashnik2021styleclip}.

\subsubsection{More qualitative comparative analysis}
\begin{figure*}
    \centering
    \includegraphics[width=\textwidth]{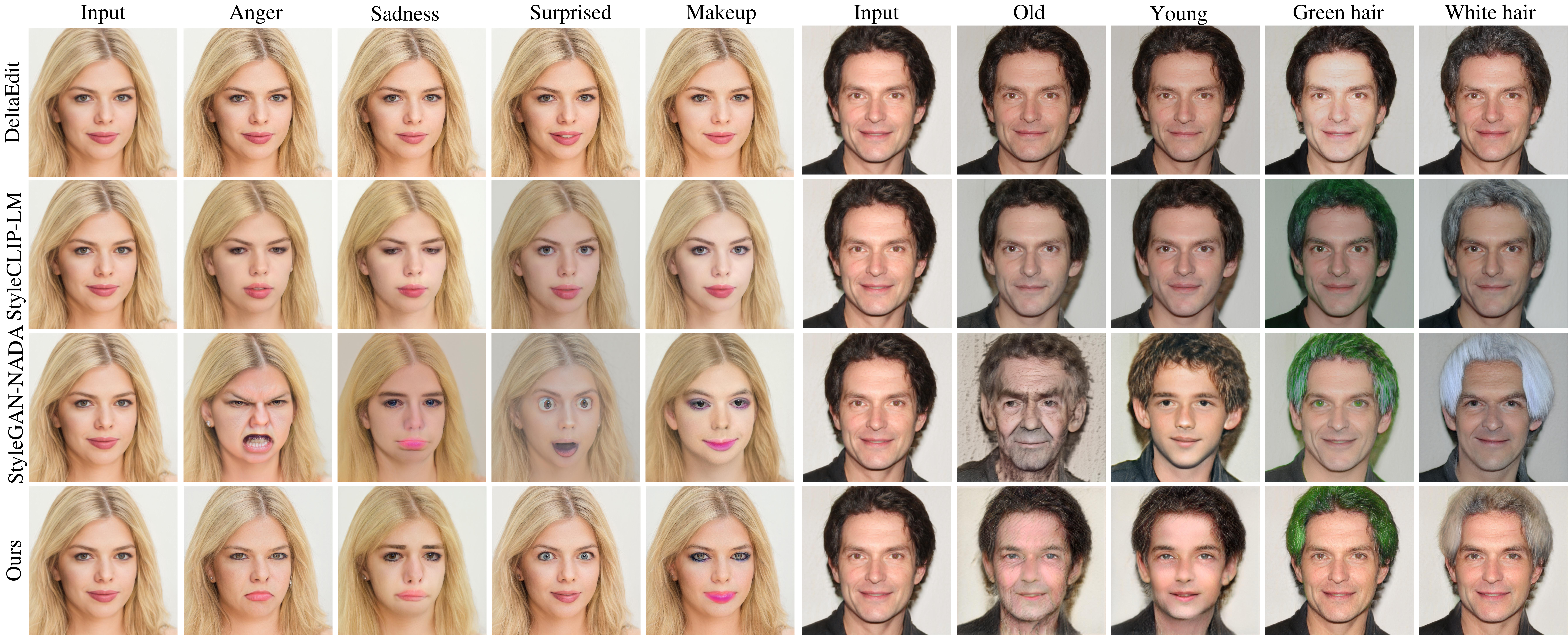}
    \caption{Qualitative comparison results of our method with DeltaEdit, StyleGAN-NADA, and StylecliP-LM. DeltaEdit could not perform precise editing operations based on the given text prompt. StyleCLIP-LM struggled to accurately edit specific text prompts (e.g., Anger, sadness, old, etc.) and affected other non-relevant image areas (e.g., surprised, Green hair, etc.). The results produced by StyleGAN-NADA have significant variations with the input face image, and the identity consistency before and after editing is severely impaired. In contrast, our method successfully edited the image with precision and without altering the non-relevant regions of the image.}
    \label{fig:suplement_compare_eidt}
\end{figure*}

\begin{figure*}
    \centering
    \includegraphics[width=0.75\textwidth]{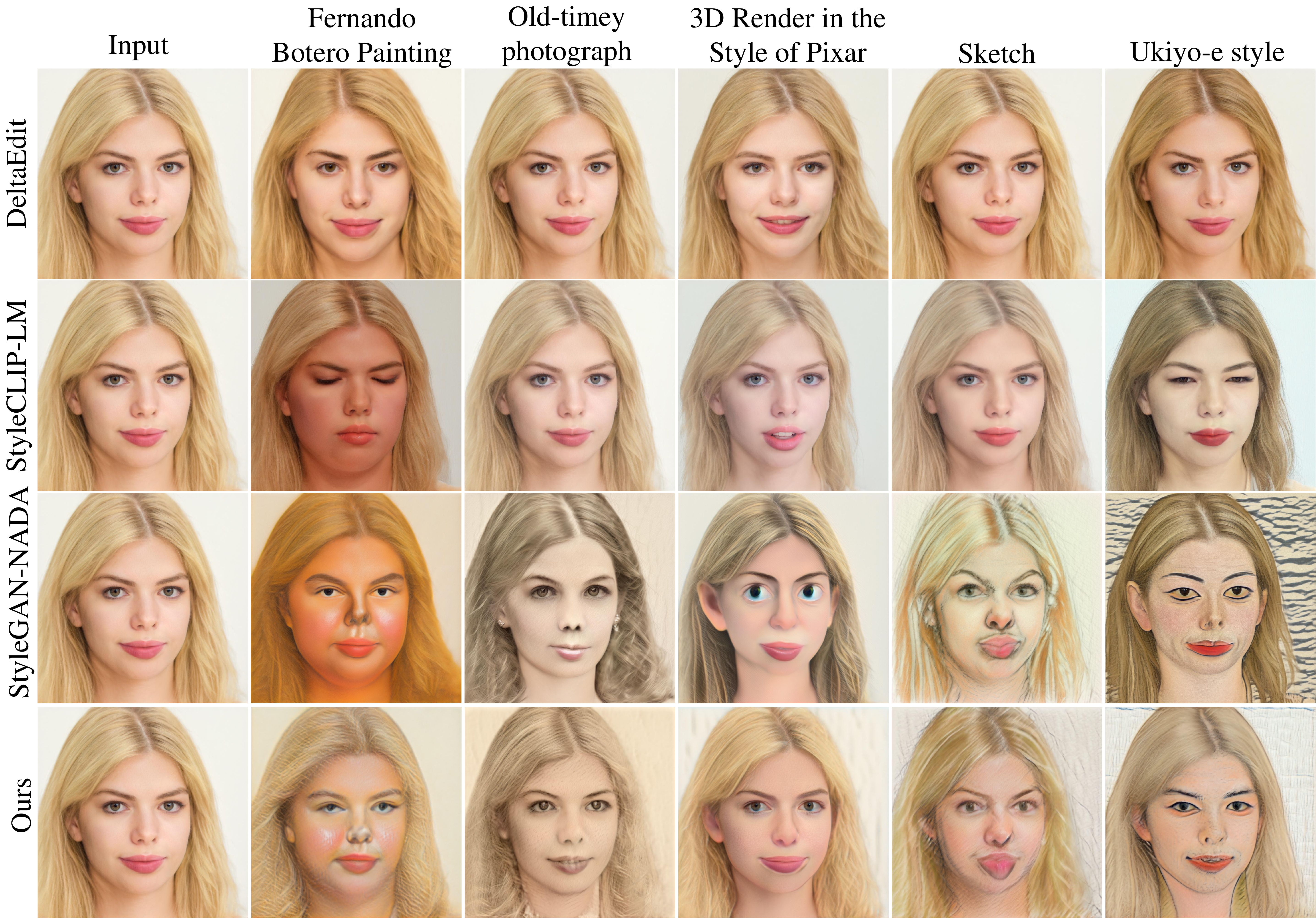}
    \caption{Qualitative comparison results of our method with DeltaEdit and StylecliP-LM. When conducting cross-domain editing, our method can achieve style transfer well, while the method based on manipulating latent codes cannot effectively transfer the style of the image. Although StyleGAN-NADA performs well on style transfer, our method has better identity protection ability while achieving cross-domain editing.}
    \label{fig:style_compare}
\end{figure*}

\begin{figure*}
    \centering
    \includegraphics[width=\textwidth]{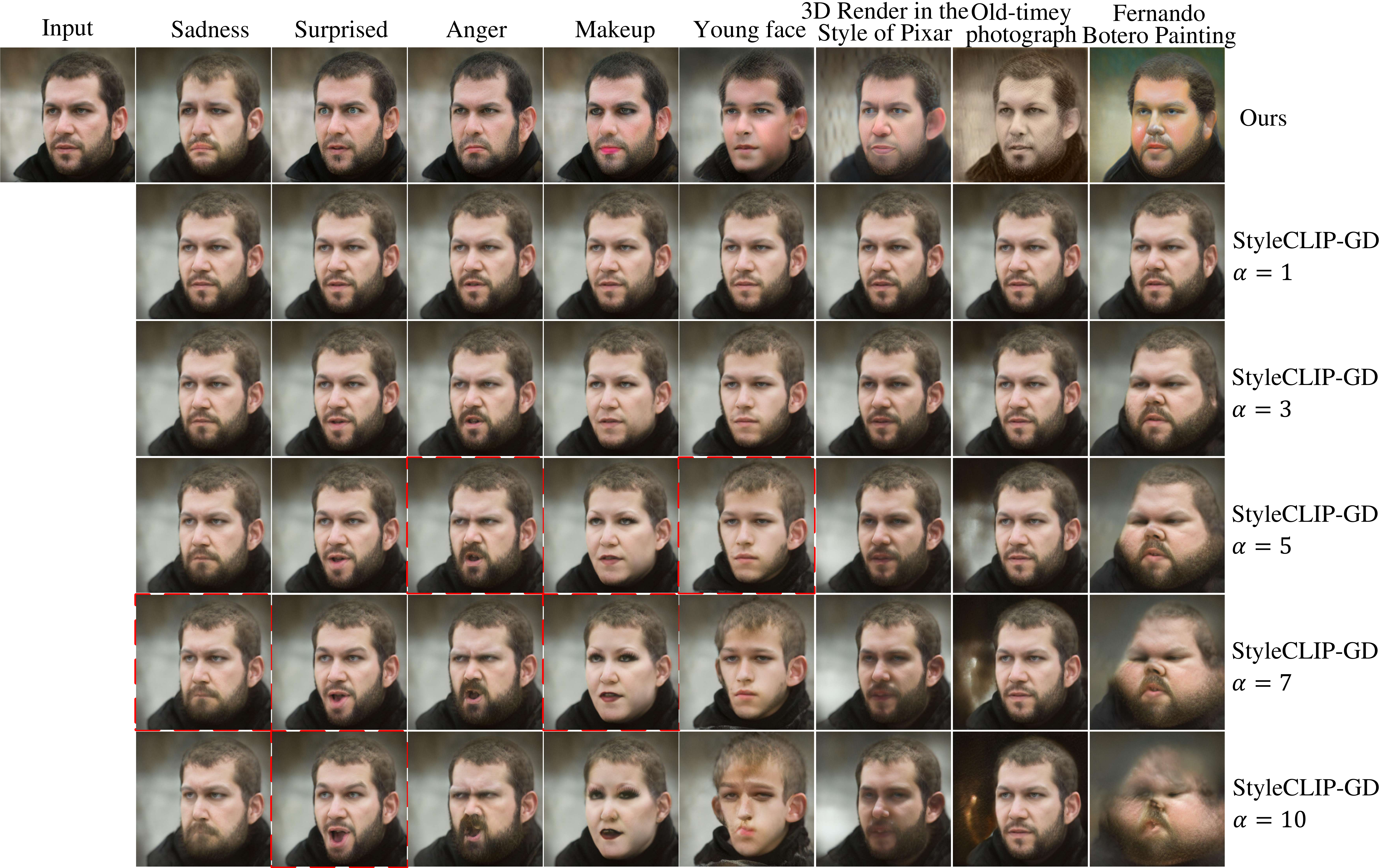}
    \caption{Qualitative comparison results between our and the StyleCLIP-GD methods with different parameter Settings. The red dashed lines indicate the edited images considered best by StyleCLIP-GD with various parameter settings based on visual inspection. Our method surpasses StyleCLIP-GD in image editing without parameter tuning (e.g., Sadness, makeup). We can also perform cross-domain style transfer, which SyleCLIP-GD cannot.}
    \label{fig:styleclip}
\end{figure*}

In Figure \ref{fig:suplement_compare_eidt} and Figure \ref{fig:style_compare}, we present a more detailed qualitative comparison of our method with StyleCLIP-LM, StyleGAN-NADA, and DeltaEdit. The results demonstrate that our method outperforms the others, as DeltaEdit fails to edit the images according to the target texts accurately. StyleCLIP-LM achieves accurate editing of the target region but causes adverse effects on non-target regions. Most importantly, our method can achieve cross-domain style transfer, which cannot be achieved by these methods based on manipulating latent codes. Compared with StyleGAN-NADA, our method performs better on the controllability of attribute changes and the consistency of images before and after editing. Additionally, we compare the StyleCLIP-GD method with different parameter settings and visualize the results in Figure \ref{fig:styleclip}. Our method achieves more accurate image editing without the need for any parameter tuning based on text conditions.

\subsubsection{User study}
To assess the visual realism of the editing performance, we conducted a user study involving 20 participants who completed 20 rounds of evaluation, resulting in a total of 400 votes. Each round of evaluation shows different editing targets, including 15 facial attributes and five style attributes in total. Each vote recorded the most accurately edited image in each evaluation round. The percentage of votes for each method has been summarized in Table \ref{tab:user}. The results indicate that our approach is favored by the majority of participants in terms of editing accuracy.

\begin{table}[htbp]
  \centering
  \caption{The users evaluated the results of images generated by different methods, where ACC represents the image that achieved the most accurate editing effect according to the target text prompt. The values in the table represent the percentage of results the user selects.}
    \begin{tabular}{l|p{1.5cm}<{\centering}}
    \toprule
    Methods & ACC ↑   \\
    \midrule
    StyleCLIP-LM &    13.00\%    \\
    StyleCLIP-GD $\alpha=5 $  &   8.75\%      \\
    StyleCLIP-GD $\alpha=10 $   & 3.75\%      \\
    DeltaEdit &     7.75\%    \\
    StyleGAN-NADA &   15.00\%      \\
    Ours  &    \textbf{51.75}\%    \\
    \bottomrule
    \end{tabular}%
  \label{tab:user}%
\end{table}%

\subsection{Extended ablations studies}
\begin{figure*}
    \centering
    \includegraphics[width=0.8\textwidth]{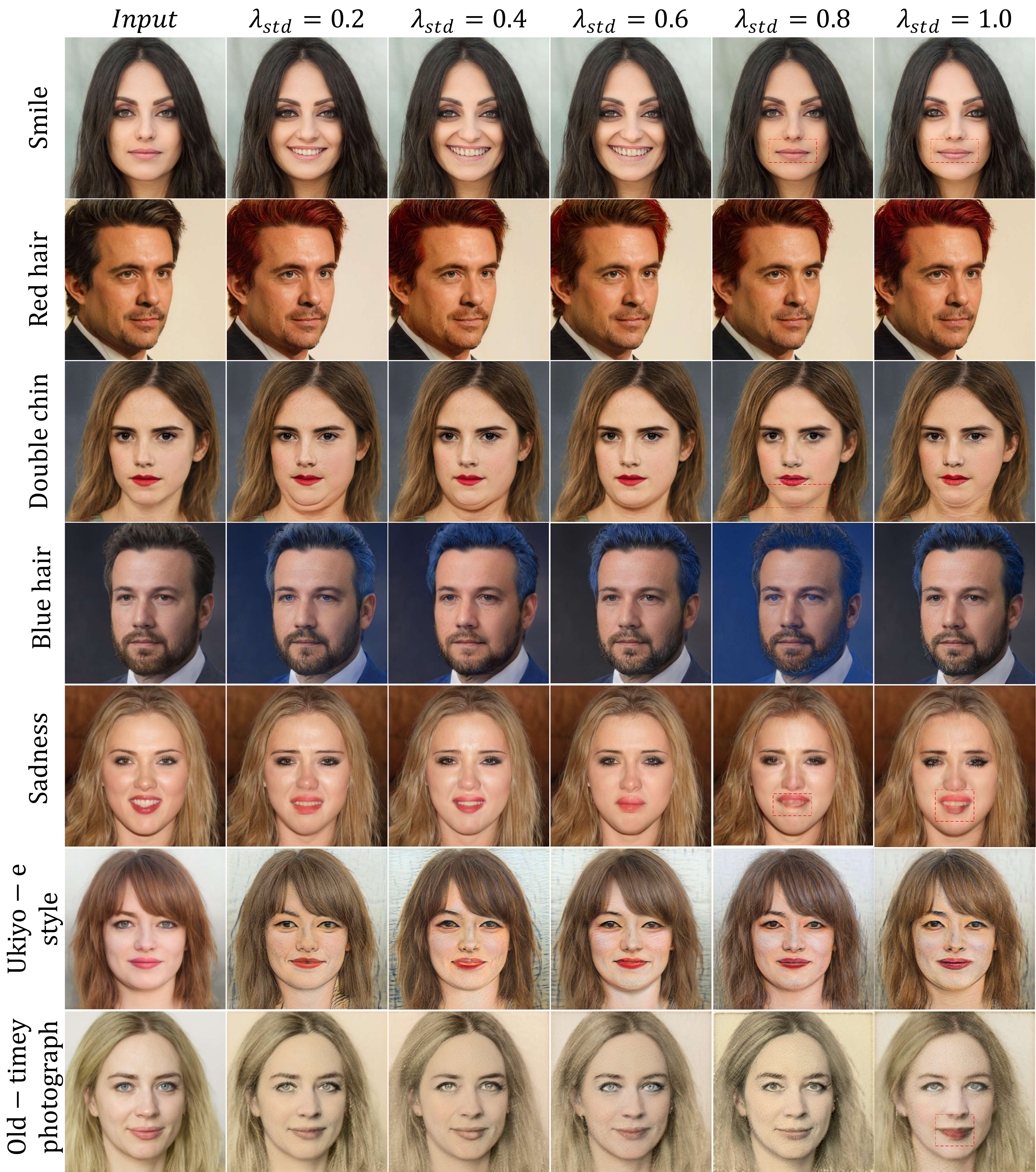}
    \caption{Results generated by our method with different values of $\lambda_{std}$ when using adaptive layer selector. When $\lambda_{std} \geq 0.8$, the editing effect of specific attributes (e.g., smile, sadness) starts to deteriorate.}
    \label{fig:delta_choose}
\end{figure*}

In selecting the adaptive threshold $\varphi$ in the adaptive layer selector, the hyperparameter $\lambda_{std}$ is crucial. To determine the appropriate hyperparameter $\lambda_{std}$, we conducted different qualitative and quantitative analyses for $\lambda_{std}=0.2$, $\lambda_{std}=0.4$, $\lambda_{std}=0.6$, $\lambda_{std}=0.8$, and $\lambda_{std}=1.0$, respectively. As shown in Figure \ref{fig:delta_choose}, we observed that when the value of $\lambda_{std}$ is too high, the results of image editing become worse or even have no effect (e.g., when we edit the “smile” attribute, the image remains unchanged when $\lambda_{std}\geq0.8$). This is because a higher $\lambda_{std}$ value results in fewer layers that output weight factors being selected by the adaptive layer selector.

We provide objective indicators such as PSNR, LPIPS, SSIM, IDS, and CS for the six image changes before and after in Figure \ref{fig:delta_choose}, respectively. In Table \ref{tab:addlabel}, we list the rankings of different $\lambda_{std}$ values for each indicator and calculate the average ranking (AR) to select the most appropriate $\lambda_{std}$. We observe that $\lambda_{std}=0.6$ and $\lambda_{std}=1.0$ have the same average rank (AR=2), but our main objective is to achieve more authentic image editing. Thus, we choose $\lambda_{std}=0.6$, which has a higher CS value. Additionally, when $\lambda_{std}=1.0$, fewer changes are generated before and after image editing, leading to higher values in indicators like PSNR, SSIM, and IDS.

\subsection{Limitation analysis}

\begin{figure}
    \centering
    \includegraphics[width=\linewidth]{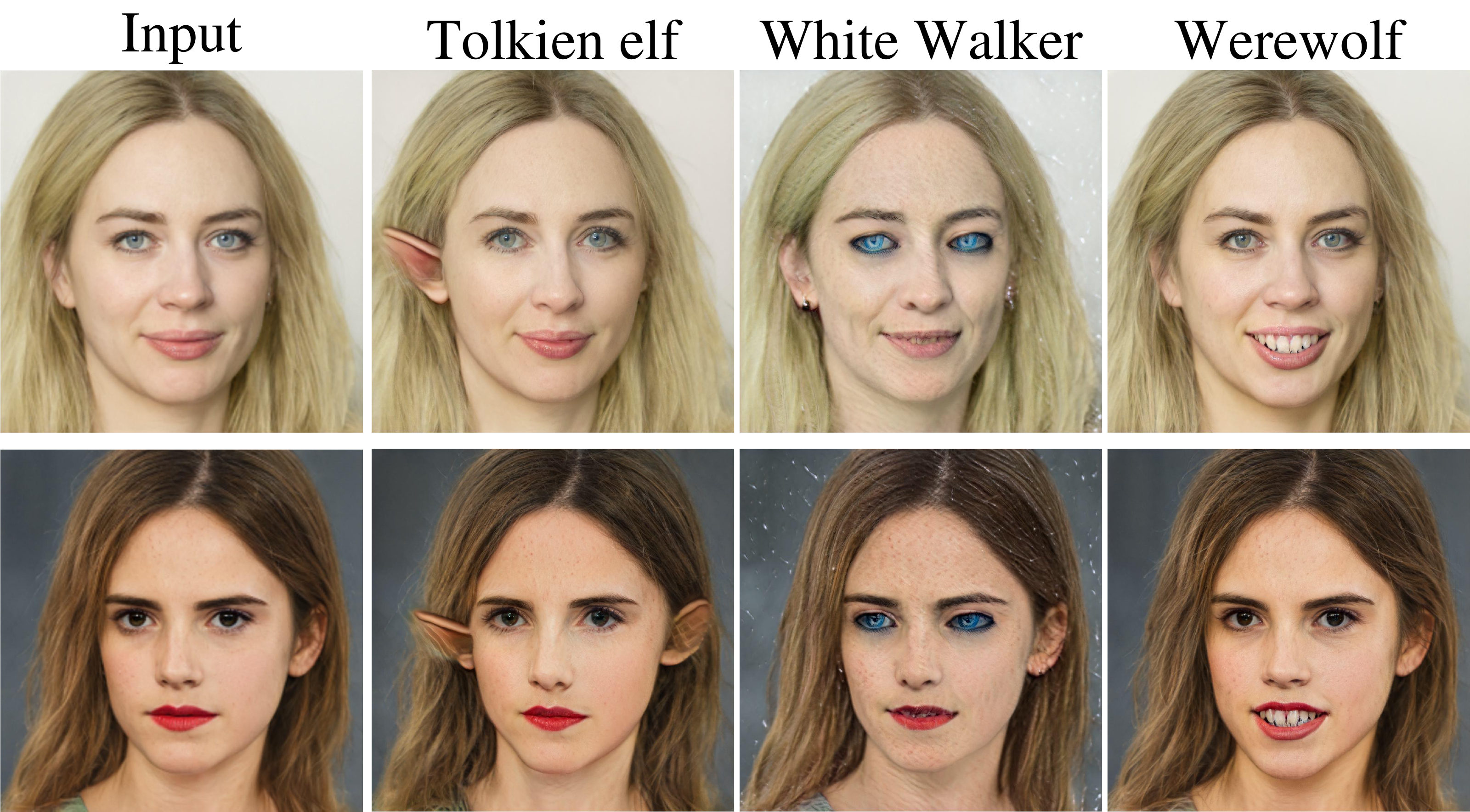}
    \caption{The result of editing the original image to the target domain image using our method. The target style described in the text prompt is displayed above the image.}
    \label{fig:limits}
\end{figure}

Our method excels in authentically editing various attributes of real images and in editing real images to other image domains that are not present during the entire training process. However, when we apply our method to edit an image to a specific image domain, as illustrated in Figure \ref{fig:limits}, only some of the properties of the real image are modified, and the image does not fully transform to the target style, we analyze that our method is based on the reassignment of the generator's weights, which is different from the retraining approach. The latter directly adjusts the weights through fully supervised learning, while our method achieves weight changes through scaling adjustments. Despite multiple training iterations, obtaining the optimal weight factor to scale the original generator weights to the optimal solution ideally is challenging. Nevertheless, our method retains an excellent image editing ability compared to the retraining of generator weights. In future work, we will further explore broader image editing tasks, including attribute editing, style transfer, and other related jobs, using a single network architecture.

\begin{table*}[htbp]
  \centering
  \caption{Results of $\lambda_{std}$ for adaptive layer selector threshold values $\varphi$. The numbers in parentheses represent the ranking of the corresponding values in the comparison. 'AR' indicates the average rank, and 'Nop' means the number of parameters in hypernetworks, measured in millions.}
  
    \begin{tabular}{c|p{2cm}<{\centering}p{2cm}<{\centering}p{2cm}<{\centering}p{2cm}<{\centering}p{2cm}<{\centering}p{1cm}<{\centering}p{2cm}<{\centering}}
    \toprule
          & PSNR↑  & LPIPS↓  & SSIM↑  & IDS↑   & CS↑    & AR↓  & Nop↓\\
    \midrule
    $\lambda_{std}=0.2$   & 25.06(4) & 0.20(2) & 0.86(2) & 0.83(3) & 26.33(3) & 2.8 &27.24M\\
    $\lambda_{std}=0.4$   & 24.60(5) & 0.21(3) & 0.85(3) & 0.78(5) & 27.05(2) & 3.6 &23.54M\\
    $\lambda_{std}=0.6$   & 25.75(3) & 0.20(2) & 0.86(2) & 0.84(2) & 27.12(1) & 2 &16.28M\\
    $\lambda_{std}=0.8$   & 26.89(1) & 0.20(2) & 0.86(2) & 0.82(4) & 25.55(4) & 2.6 &12.01M\\
    $\lambda_{std}=1.0$     & 26.79(2) & 0.19(1) & 0.88(1) & 0.85(1) & 25.20(5) & 2 &11.09M\\
    \bottomrule
    \end{tabular}%
  \label{tab:addlabel}%
\end{table*}%

\end{document}